\documentclass[preprint,5p,times,twocolumn,a4paper]{elsarticle}
\usepackage{blindtext}
\usepackage[T1]{fontenc}
\usepackage[utf8]{inputenc}
\usepackage{url}
\usepackage{subfig}
\usepackage{amsmath,amssymb,amsfonts}
\usepackage[toc]{appendix}
\usepackage{algorithmic}
\usepackage{graphicx}
\usepackage{textcomp}
\usepackage{adjustbox}
\usepackage{graphicx}
\usepackage{caption}
\usepackage[textsize=tiny,textwidth=1.4cm]{todonotes}
\usepackage{mathtools,lipsum, nccmath,cuted}
\usepackage{siunitx}
\usepackage{textcomp}

\usepackage{comment}
\usepackage{makecell}
\usepackage{algorithm}
\usepackage{algorithmic}
\usepackage{gensymb}
\usepackage{xcolor}
\definecolor{Gray}{gray}{0.925}
\usepackage{colortbl}
\usepackage{tabularx}
\usepackage{array}
\usepackage{rotating}
\usepackage{pdflscape}
\usepackage[normalem]{ulem}
\usepackage{titlesec}
\usepackage{soul}
\usepackage{url}
\usepackage{natbib}
\usepackage{balance}
\usepackage[capitalise]{cleveref}

\usepackage{enumitem}
\graphicspath{ {figures/} }
\def\BibTeX{{\rm B\kern-.05em{\sc i\kern-.025em b}\kern-.08em
    T\kern-.1667em\lower.7ex\hbox{E}\kern-.125emX}}
    
\usepackage{pgfplots}
\usepgfplotslibrary{colorbrewer}
\pgfplotsset{compat = 1.14, cycle list/Set1-8} 
\usetikzlibrary{pgfplots.statistics, pgfplots.colorbrewer,pgfplots.dateplot,
        shadows,
        matrix} 

\usepackage{pgfplotstable}
\usepackage{filecontents}
\usepackage{graphicx}
\pgfplotsset{compat=1.14}

\definecolor{blueLine}{RGB}{57,106,177}
\definecolor{blueFill}{RGB}{114,147,203}
\definecolor{redLine}{RGB}{204,37,41}
\definecolor{greenline}{RGB}{0,250,0}
\definecolor{blackLine}{RGB}{0,0,0}
\definecolor{goldLine}{RGB}{160,82,45}

\usepackage[font=footnotesize]{caption}
\usepackage{soul}

\usepackage{acro}

\DeclareAcronym{cps}{
  short = CPS,
  long  = Cyber Physical Systems,
  sort  = C,
}
\DeclareAcronym{lec}{
  short = LEC,
  long  = Learning Enabled Component,
  sort  = L,
}
\DeclareAcronym{cnn}{
  short = CNN,
  long  = Convolutional Neural Network,
  sort  = C,
}

\DeclareAcronym{dnn}{
  short = DNN,
  long  = Deep Neural Network,
  sort  = D,
}

\DeclareAcronym{rl}{
  short = RL,
  long  = Reinforcement Learning,
  sort  = R,
}

\DeclareAcronym{ld}{
  short = LD,
  long  = Lane Detection,
  sort  = l,
}

\DeclareAcronym{e2e}{
  short = e2e,
  long  = end-to-end,
  sort  = e,
}

\usepackage[font=footnotesize]{caption}

\journal{Special Issue on the 2019 IEEE Symposium on Real-time Computing ISORC}

\begin{document}

%\section*{empty. Please skip to next page}
%%\newpage
%\setcounter{page}{0}

\begin{frontmatter}

\title{Dynamic-Weighted Simplex Strategy for Learning Enabled Cyber Physical Systems}

% \title{Mechanisms for Run-time Safety Supervision in Resource Constrained Learning Enabled Systems}

\author{Shreyas Ramakrishna} 

\author{Charles Harstell}

\author{Matthew P Burruss}

\author{Gabor Karsai}

\author{Abhishek Dubey}

\address{\textit{Institute for Software Integrated Systems}\\
\textit{Vanderbilt University}
\\ Nashville, TN, USA
}

\begin{abstract}
Cyber Physical Systems (CPS) have increasingly started using Learning Enabled Components (LECs) for performing perception-based control tasks. The simple design approach, and their capability to continuously learn has led to their widespread use in different autonomous applications. Despite their simplicity and impressive capabilities, these models are difficult to assure, which makes their use challenging. The problem of assuring CPS with untrusted controllers has been achieved using the Simplex Architecture. This architecture integrates the system to be assured with a safe controller and provides a decision logic to switch between the decisions of these controllers. However, the key challenges in using the Simplex Architecture are: (1) designing an effective decision logic, and (2) sudden transitions between controller decisions lead to inconsistent system performance. To address these research challenges, we make three key contributions: (1) \textit{dynamic-weighted simplex strategy} -- we introduce ``weighted simplex strategy" as the weighted ensemble extension of the classical Simplex Architecture. We then provide a reinforcement learning based mechanism to find dynamic ensemble weights, (2) \textit{middleware framework} -- we design a framework that allows the use of the dynamic-weighted simplex strategy, and provides a resource manager to monitor the computational resources, and (3) \textit{hardware testbed} -- we design a remote-controlled car testbed called DeepNNCar to test and demonstrate the aforementioned key concepts. Using the hardware, we show that the dynamic-weighted simplex strategy has 60\% fewer out-of-track occurrences (soft constraint violations), while demonstrating higher optimized speed (performance) of 0.4 m/s during indoor driving than the original LEC driven system.
\end{abstract}


\end{frontmatter}

{\footnotesize
\printacronyms[name=Abbreviations]
}

\section{Introduction}
\label{sec:introduction}

Cyber Physical Systems (CPS) have increasingly started relying on the use of Learning Enabled Components (LECs) as part of the control loop, performing varied perception-based autonomy tasks. These data-driven components are trained using machine learning approaches like deep learning \cite{goodfellow2016deep}, and reinforcement learning (RL) \cite{sutton2018reinforcement}. These approaches have given these components the capability to continuously learn and work in unfamiliar environments. Recently, these components are seeing widespread acceptance and use in various autonomous applications like NVIDIA's DAVE-II \cite{bojarski2016end} convolutional neural network (CNN) that performs \ac{e2e} learning-based self-driving, and Tesla's autonomous driving that has recently completed 2 billion driving miles \cite{teslamiles} using several LECs to perform object detection and tracking, and image segmentation \cite{badrinarayanan2017segnet}.

% : NVIDIA \cite{bojarski2016end} used their DAVE-II convolutional neural network (CNN) model to perform \ac{e2e} learning-based self-driving. Tesla \cite{teslamiles} has recently completed two billion driving miles using autonomous driving that uses several of these components to perform object detection and tracking, image segmentation \cite{badrinarayanan2017segnet}, etc.   

Despite their impressive capabilities, it is still a challenge to assure the correctness of these systems under all circumstances. Applying conventional assurance based development \cite{graydon2007assurance} or designing safety and assurance cases \cite{bloomfield2010safety} is complicated for these systems because of the following limitations: First, the black-box nature of these models limits testing coverage. Pei, Kexin, et al. \cite{pei2017deepxplore} discuss the limitations of performing exhaustive testing of these models, and introduce a whitebox framework called DeepXplore that can perform limited testing. Secondly, the existing verification tools are limited by the complexity of the neural network and non-linearity of the activation functions. The authors in \cite{xiang2017reachable} discuss the limitations of performing verification techniques on CNNs. Third, these components learn from data, and they perform well only when the test observations resemble the training data. The authors in \cite{pei2017deepxplore}, \cite{boloor2019attacking} have shown how subtle changes (or adversaries) in the test image confuse the model and results in erroneous predictions. These limitations make it challenging to design safety or assurance cases for these systems. Recently, the DARPA Assured Autonomy project \cite{darpa} has been focusing on designing tools to overcome these limitations.

\begin{figure}[t]
\centering
\setlength{\belowcaptionskip}{-10pt}
\setlength{\abovecaptionskip}{-4pt}
 \includegraphics[width=0.8\columnwidth]{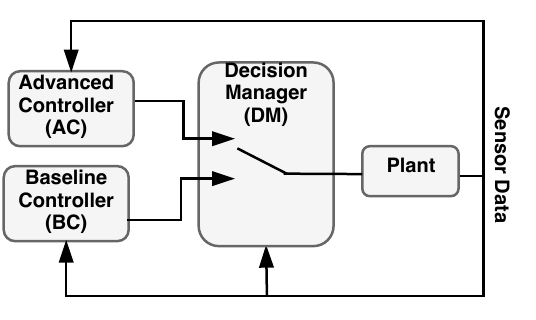}
 \caption{The Simplex Architecture adopted from \cite{bak2009system} combines an unverified advanced controller (AC) with a safe baseline controller (BC) and a decision manager (DM). The DM is responsible for selecting the controller that can maintain the system safe.}
 \label{fig:simplex}
 \vspace{-0.4em}
 \end{figure}

The problem of assuring safety guarantees in CPS has been tackled using the Simplex Architecture \cite{seto1998simplex}. This architecture (see \cref{fig:simplex}) improves the system's safety by combining an unverified high-performance controller (Advanced Controller (AC)) with a safety controller (Baseline Controller (BC)) and a decision manager (DM). The DM is the core component in the Simplex Architecture, which arbitrates the control between the two controllers based on a safety criteria. This architecture requires verifying only the BC and the DM, thus alleviating the requirement of verifying the AC (e.g. LEC), which is sometimes difficult or not possible. Simplex Architectures have been used in several applications: the authors in \cite{crenshaw2007simplex} have demonstrated the utility of the Simplex Architecture to avoid a collision in a fleet of remote-controlled cars. \cite{seto2000case} discusses a case study of using the Simplex Architecture for the automatic landing of a F-16 aircraft. These applications demonstrate the use of Simplex Architecture for complex and safety-critical CPS applications.

% \textcolor{red}{make this figure tigher - too much space is being wasted.}
 
% has a decision manager (DM) to arbitrate between the advanced controller (AC) and the baseline controller (BC). The arbitration of the control happens when the control actions of the AC starts moving the system towards unsafe states. }

However, the key challenges in using the Simplex Architectures for CPS applications are: (1) \ul{designing effective decision logic}: as discussed in \cite{bak2014real}, the critical challenge is the design of appropriate decision logic. It is trivial to design safe logic by always using the BC to make the decisions. However, that  would make the system too defensive without utilizing the high performance AC. Therefore, the the architecture should seek to utilize the AC as much as possible to avoid conservatively using the BC. That is, it is important to design decision logic that balances the safety and performance of the system. Key challenge (2) \ul{sudden transitions}: the instantaneous transitions (see \cref{fig:wss}) from the AC to the BC drastically degrades the system performance. For example, Bak, Stanley, et al.\cite{bak2009system} have discussed the applicability of the Simplex Architecture in heart pacemakers. The authors discuss how sudden transitions between simplex controllers are dangerous in safety critical applications. For a pacemaker case study, they illustrate how sudden jumps between two discrete heart rates (65 to 120 beats per minute) from two simplex controllers can make the patients dizzy and uncomfortable. Similarly, in automotive applications the sudden changes in the control (e.g. speed) will be perceived as jerks (high rate of change of acceleration) and can be uncomfortable for passengers (e.g. Toyota sudden acceleration problem \cite{finch2009toyota}). 
%As illustrated, such sudden transitions are not good for safety-critical CPS applications and have to be avoided.

% They have further discussed the importance of performing smooth transitions among the simplex controllers in such sensitive safety-critical applications. To illustrate, they have also discussed how sudden jumping between two discrete heart rates (65 to 120 beats per minute) of the two simplex controllers could make the patients dizzy and uncomfortable. Similarly, in autonomous applications this sudden change will be perceived as a jerk (high rate of change of acceleration) and will be uncomfortable to passengers.

% The red line shows a very smooth curve, however the granularity of the weights decides the smoothness of the transition. The curve may have more of a step transition.

\textbf{Contributions:} To address these challenges
%key challenges in designing decision logic and performing smoother transitions between simplex controllers, especially for systems with LECs, 
we perform a weighted blending of the simplex control actions. We refer to this weighted extension of the conventional Simplex Architecture as the "weighted simplex strategy". Instead of selecting a single controller action, this strategy computes a weighted ensemble of all the controllers actions. Such blending mechanisms have shown to improve performance or accuracy in model ensembles \cite{jimenez1998dynamically}, and multiple model adaptive predictive control \cite{zhang2012stable}. For the Simplex Architecture, we hypothesize blending the controllers actions could optimally balance the performance and soft constraint violations of the system while avoiding abrupt transitions. Specifically, this strategy aims at optimizing the performance while reducing the constraint violations. So, it can only be used for systems which can tolerate soft constraint violations (ie. constraints that can be violated, but will incur a penalty). We summarize our contributions below.

% This strategy provides a higher preference to the safety of the system, but as it , the strategy is currently applicable to systems which can tolerate soft constraint violations (constraints that can be violated, but will incur a penalty). We summarize our contributions below.

% Note that this approach is suitable for systems which can tolerate soft constraint violations (constraints that can be violated, but will incur a penalty), for example LEC driven autonomous systems that are used in hospitals \cite{ozkil2009service}, warehouse \cite{bhasin2016amazon}, and laboratory research testbeds \cite{F1/10}, \cite{karaman2017project}.
%We hypothesize that the weights used for blending can be controlled to optimally balance the performance and soft constraint violations of the system while avoiding abrupt transitions. This blending strategy is mostly suitable for systems which can tolerate soft constraint violations (constraints that can be violated, but will incur a penalty). Furthermore, the key challenge in using the weighted simplex strategy is to design a mechanism to find appropriate dynamic ensemble weights according to the context of the application. To find these dynamic weights we introduce the dynamic-weighted simplex strategy. 

\begin{itemize}[noitemsep,leftmargin=*]
    \item \ul{Dynamic-Weighted Simplex Strategy} -- We discuss a mechanism to find dynamic ensemble weights of the weighted simplex strategy using reinforcement learning (RL). We show the design of the reward function that is responsible for reducing the soft constraint violations while improving the performance of the system. 
    
    \item \ul{Middleware Framework} -- We design a middleware framework to deploy the weighted simplex strategy on a physical CPS platform called DeepNNCar. In addition, the framework also has a resource manager that is used to monitor the computational resource of the system while mitigating any overload introduced by the complex computations of the proposed strategy.
    
    \item \ul{Hardware Testbed} -- We discuss the design of a resource constrained remote controlled car, called the DeepNNCar, that will be used as a case study to test the dynamic weighted blending mechanism and the middleware framework. 
\end{itemize}

%The proposed framework provides a robust, lightweight mechanism for runtime supervision of resource-constrained LEC driven autonomous systems that are used in hospitals \cite{ozkil2009service}, warehouse \cite{bhasin2016amazon}, and laboratory research testbeds \cite{F1/10}, \cite{karaman2017project}. 

%This paper is an extension of our previous work \cite{ramakrishna2019augmenting} and provides detailed discussion on the dynamic-weighted simplex strategy. 
%that was submitted to ISORC 2019. Here, we explain in greater detail the dynamic-weighted simplex strategy and describe the RL setup to compute the dynamic ensemble weights. 
%We have also extended the resource management section with a Deep Neural Network (DNN) forecasting model to predict the next state system utilization (e.g. CPU utilization and temperature) of the onboard computational unit (RPi3).  

\begin{table}[t]
    \centering
    \caption{Notation Lookup Map}
    \setlength{\belowcaptionskip}{-6pt}
    \renewcommand{\arraystretch}{1.2}
    \footnotesize
    \begin{tabular}{|c|p{7cm}|}
    \hline
    Symbol& Description\\
    \hline
    $\theta_L$ & Steering PWM value of DeepNNCar using LEC controller\\
    \hline
    $\theta_{C}$ & Steering PWM value of DeepNNCar using OpenCV controller\\
    \hline
     $\theta_{D}$ & Steering PWM value using dynamic-weighted simplex strategy\\
     \hline
    $\theta_{F}$ & Steering PWM value using fixed-weighted simplex strategy\\
    \hline
    $W_L$ &  Ensemble weight given to LEC controller\\
    \hline
    $W_{C}$ & Ensemble weight given to OpenCV controller\\
    \hline
    $W_{SET}$ & Ensemble weights \{$W_L, W_{C}$\} computed by dynamic-weighted simplex strategy\\
    \hline
    $T_{R}$ & Inference pipeline time of DeepNNCar\\
    \hline
    $\nu_t$ & Current speed PWM value of DeepNNCar\\
    \hline
    \end{tabular}
    \label{Table:SymbolTable}
    \vspace{-0.4em}
\end{table}

% to use a neural network based temperature forecasting model that predicts the next temperature of the onboard computational unit (RPi3). 

\textbf{Outline}: The outline of the this paper is as follows: in \cref{sec:Back}, we discuss the background topics. In \cref{sec:RW}, we present related research. \cref{sec:deepnncar} describes our DeepNNCar testbed and its control algorithms. \cref{sec:Simplex Strategies} introduces the dynamic-weighted simplex strategy and the setup to compute dynamic weights. In \cref{sec:RM}, we design the resource manager for managing the systems resource utilization. \cref{sec:middleware} describes the system integration. \cref{sec:evaluation} evaluates the dynamic-weighted simplex strategy and the resource manager. \cref{sec:conclusion} presents our conclusions. The notations used in the paper are described in \cref{Table:SymbolTable}.

\section{Background}
\label{sec:Back}
In this section we discuss a few key concepts that are required to understand our methodology discussed in the later sections of this paper. Readers who are familiar can skip this section.

\begin{figure}[t]
\centering
\setlength{\belowcaptionskip}{-10pt}
 \includegraphics[width=\columnwidth]{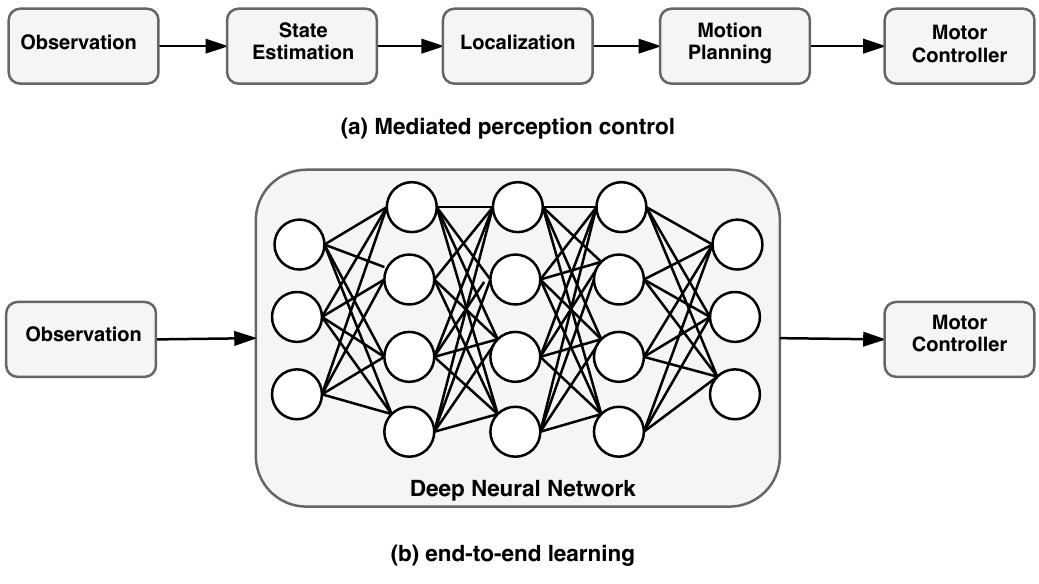}
 \caption{Perception based control approaches in robotics. Adopted from \cite{approach}.}
 \label{fig:control-approaches}
 \vspace{-0.4em}
 \end{figure}
 
% \textcolor{red}{all other figures in paper should be same shade}

\subsection{LEC Based Control Approaches in Autonomous Systems}
Perception based control paradigms in autonomous vehicles can be classified into the mediated perception approach, and end-to-end (e2e) learning approach as shown in \cref{fig:control-approaches}. 

The mediated perception approach \cite{ullman1980against}, \cite{chen2015deepdriving} decomposes the problem into multiple sub-goals which then together form a processing pipeline for performing autonomous driving. The pipeline (see \cref{fig:control-approaches}) has multiple stages of operation like sensing, state estimation, localization, motion planning and motor control, which use sensor data to learn high level representations of lanes, objects, cars, and traffic lights. This high-level information is then used by the controller to compute the low level actuation of the car.

The advantages of using this approach (as discussed in \cite{mantegazza2018vision}) are: (1) transparent internal modes of operation for testing and debugging, (2) robust decision-making capabilities due to multiple dedicated algorithms, and (3) a high degree of freedom for the designer to select and fine tune the internal stage algorithms. However, multiple stages of operation along with the requirement to create and maintain a large code base even for simple navigation tasks makes its application on small-scale systems challenging. 

In contrast, e2e learning \cite{levine2016end} is a perception-based control approach that uses supervised learning \cite{zhu2009introduction} to directly compute the control action. It has been applied to different indoor and outdoor navigation tasks, such as obstacle avoidance \cite{muller2006off}, off-road autonomous navigation systems \cite{bajracharya2009autonomous}, and autonomous driving \cite{bojarski2016end},\cite{pomerleau1989alvinn}. In this work, we use an e2e learning approach for navigating a DeepNNCar (explained in \cref{sec:deepnncar}) around an indoor track. 
In this approach, a data-driven model (e.g. DNN or CNN) takes in observations from different sensors such as camera, IMU, LIDAR, etc. and predicts an output control action that could either be the steering or speed of the system. Unlike traditional software where the execution logic is derived from analytical models, DNNs learn the underlying relationship between the observed input and the output from the data. The functionality of DNNs can be tuned with training hyperparameters like learning rates, epochs, optimization algorithms, etc. This approach has recently gained considerable attention because of its conceptual simplicity and computational efficiency.

\ul{Problems}: Some limitations of the e2e learning approach are: (1) the direct sensor-actuation mapping is only limited to perform point based predictions, which can limit the use of the approach in complex navigation based tasks, (2) LECs used in mapping the sensor to actuation values learn from training data. So, even slight distribution shift in the test observations will result in inconsistent behavior in these models, and (3) training the LEC is hard as there are several hyperparameters to be tuned. Improper tuning of these parameters can lead to underfitting or overfitting problems.

% The advanced controller (AC) is usually a high performance, but unverified, and the baseline controller (BC) is of lower performance but provides higher safety guarantees. The architecture is also equipped with a decision manager (DM) that selects one of the controller's action based on a decision logic.

\subsection{Simplex Architecture}
As discussed by Seto, et al. \cite{seto1998simplex}, the Simplex Architecture is built on the concept of analytic redundancy, i.e. redundant components with different design and implementations, but similar interfaces. \cref{fig:simplex}
illustrates the structure of the Simplex Architecture, where the analytic redundancy is achieved using two controllers with same interfaces but with different implementation and specifications. An advanced controller (AC) with high-performance specification is combined with a reliable and safe baseline controller (BC), and a decision manager (DM). The decision logic of the DM is encoded to guarantee the safety of the system, i.e. the logic will transfer the control from the AC to the BC, when the AC decisions start moving the system toward unsafe states. The simplex combination of controller outputs can be represented as the weighted combination of the two controller output and can be written as in \cref{Eqn:gegeralsimplex}:

\begin{equation}
 C_{SA} = W_1 \cdot C_{AC} + W_2 \cdot C_{BC}
 \label{Eqn:gegeralsimplex}
 \end{equation}
 
% The wide acceptance of this architecture stems from its capability of abstracting the requirement of verifying the AC, which is challenging for complex controllers (e.g. LECs). It requires only the BC and the DM to be verified.
 
\begin{figure}[t]
\centering
\setlength{\belowcaptionskip}{-10pt}
 \includegraphics[width=\columnwidth]{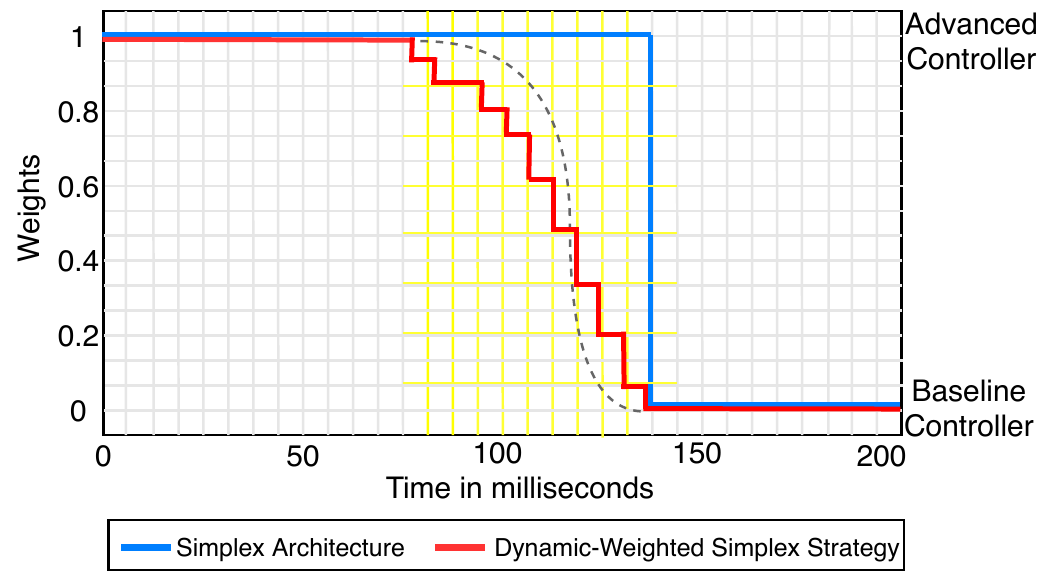}
 \caption{The weighted simplex strategy works alike the conventional simplex architecture in the high performance and safe regions. However, the targeted spectrum of this strategy is the transition region represented by yellow grid, where the unverified controller actions start drifting the system towards the unsafe state. Here, instead of instantaneous transition, the dynamic weights of the weighted simplex strategy provide a mechanism for a smoother transition between the controller outputs. The discreet weight adjustment in the dynamic-weighted simplex strategy (red line) takes a staircase function approximating a smoother curve (dotted gray line).}
 %This figure is drawn to conceptually explain the Simplex Architecture and the weighted simplex strategy.
%  \textcolor{red}{why is this a staircase - it could be any function -- make that clear.}
 \label{fig:wss}
%  \vspace{-0.4em}
 \end{figure}

Where, $C_{SA}$ is the system's control output, $C_{AC}$, $C_{BC}$ represents the controller outputs of the AC and the BC respectively. $W_1$, $W_2$ represents the weights and they sum up to 1. For the conventional Simplex Architecture, $W_1$=1 and $W_2$=0 in most scenarios, but when the AC is on the verge of jeopardizing the safety of the system, the control transfers to the BC and the weights become $W_1$=0 and $W_2$=1. The blue line in the  \cref{fig:wss} shows the control transition between the two controllers of the classical Simplex Architecture. As seen, the control remains with the AC up to a certain point, and when the DM decides to switch there is a sudden (or instantaneous) transition of the control to the BC. 

\ul{Problems}: As mentioned in \cref{sec:introduction} the key challenge of using the Simplex Architectures are: (1) designing an effective decision logic that can optimally balance the safety and performance of the system, and (2) avoiding sudden transitions in the control between the AC to BC. Such sudden control transitions may result in inconsistent system performance in some domain specific applications like pacemakers \cite{bak2009system} which are sensitive to minute changes in control actions. These challenges make it difficult to directly apply Simplex Architectures to certain sensitive domains in CPS applications.  

We solve these challenges by introducing the dynamic-weighted simplex strategy, which is the weighted extension of the classical Simplex Architecture (\cref{fig:wss}). We also introduce a mechanism to find the dynamic weights using a RL approach.

\section{Related Work}
\label{sec:RW}
This section provides an overview of the existing literature work in the dimensions of decision logic for Simplex Architecture, middleware framework for small scale CPS platforms, and autonomous robot testbeds. A complete survey of papers in these areas is beyond the scope of this paper. 

% We discuss a few relevant papers in each of these areas. A complete survey of papers in these areas is beyond the scope of this paper. 

\subsection{Decision Logic for Simplex Architectures} 
The existing literature on designing decision logic for the Simplex Architecture can be classified into two categories: 
\begin{itemize}[noitemsep,leftmargin=*]
    \item Lyapunov function-based techniques (linear matrix inequality (LMI) \cite{seto1999case}) have been used widely as the switching criteria for continuous systems. LMI is based on designing a verifiable decision logic by solving linear matrix inequalities. The authors in \cite{lee2005dependable} have used LMI tools to design the decision logic for inverted pendulum control.  
    
    \item Reachability analysis \cite{bak2011sandboxing} is another technique used popularly in hybrid systems. In this technique the systems are modeled as hybrid automata, and a set of reachable states by the system is computed. A comprehensive discussion about reachable state computations is made in \cite{bak2011sandboxing},  \cite{asarin2000approximate} .
\end{itemize}

In addition, there are a few other techniques used to design decision logic for the Simplex Architecture. The authors in \cite{bak2014real} have designed a switching criterion called Real-Time reachability algorithm. The algorithm proposes a unified approach that uses the offline LMI results along with online reachability analysis to design an effective decision logic. In \cite{vivekanandan2016simplex}, the authors have designed a simple application specific decision logic by designing a set of conditions under which the AC fails and when the controller transition must be made. This decision logic has been used in the Simplex Architecture to control an unmanned aerial vehicle (UAV). Phan, Dung et al. \cite{DBLP:conf/acsd/PhanYCGSSS17} have proposed the use of Assume-Guarantee (A-G) contracts to construct the decision logic for Simplex Architectures. A-G is a powerful reasoning, proof-based technique that show the guarantees that the system can provide when a set of assumptions are satisfied. They demonstrate their approach on a quick bot rover. 

All these discussed techniques are based on either traditional control theory, reachability analysis or pre-defined rules. The rule-based technique does not scale well across applications and would require us to write new rules for different applications. The reachability-based techniques would be too slow to implement at runtime if the state space is large. Also, the reachability analysis for LECs using DNNs is still limited to feed forward ReLU based networks \cite{lomuscio2017approach} and gets complex for CNNs. 

In this work, we use an RL algorithm as the decision logic of the simplex combination. For this we design a reward function that involves information about the performance and the constraint (safety) of the system. The motivation for using RL are as follows: (1) the inability of the traditional verification techniques to verify complex CNNs, (2) the reward based system of RL that can continuously learn to choose safe actions in dynamic environments, and (3) an RL algorithm takes very less time (typically fractions of millisecond, depending on the complexity of state space) to compute a decision, as compared to the traditional verification techniques. Such short time decision computation is critical at runtime for CPS testbeds.  

% To design a decision logic for our LEC based DeepNNCar we use RL. We design the reward function of RL to use a safety factor and a performance factor. We were encouraged to select RL because of the following reasons: (1) the inability of the traditional verification techniques to verify complex CNNs, (2) the reward based system of RL that can continuously learn to choose safe actions in dynamic environments, and (3) an RL algorithm takes very less time (typically fractions of millisecond, depending on the complexity of state space) to compute a decision, as compared to the traditional verification techniques. Such short time decision computation is critical at runtime for CPS testbeds.  

% We hypothesize the RL algorithm can dynamically learns to optimally balance the safety and performance of the system.

\subsection{Middleware Framework for Small Scale Robots}
There are several middleware frameworks designed for managing the functionalities of small-scale robots. Seiger, Ronny, et al \cite{seiger2015capability} have designed a Robot Operating System (ROS) \cite{quigley2009ros} based high-level programming framework to control the functionalities of domestic robots. ROS commands are used through the framework to control the robot in manual, semi-autonomous and fully-autonomous modes. The authors in \cite{de2016middleware} have introduced a middleware platform for distributed applications involving robots, sensors and the cloud. The framework uses AIOLOS \cite{bohez2014enabling}, which is a distributed software framework that allows the developer to design software on multiple components (sensors, robots) without having them to design the inter-component communication. This framework seeks to make the software design process easier and distributed to developers. The F1/10 platform \cite{F1/10} is a small-scale remote-controlled race car that uses a ROS based communication framework to control the steer and speed actuations of the car.

In \cite{liu2016cyber}, the authors have discussed an architecture for intelligent manufacturing units in shop floors. The key responsibility of their middleware layer is to manage devices, define interfaces and data management. The key focus of the middleware layer is to relay the data collected from the robotic sensors to the server for analysis. \cite{samad2018multi} discusses a multi-agent cloud-based framework for management of collaborative robots. The sensor data collected by a swarm of robots is sent to a cloud-based server over the internet. The server then computes the actions that need to be taken by the robots. 

All these frameworks aim at providing a middleware layer to control the actuations of the system. However, the key contribution of our middleware framework is the resource management strategy. We hypothesize an optimal resource management strategy is required to improve the performance of CPS applications.

\subsection{Autonomous Robot Testbeds} 
There have been several ongoing research projects related to physical testbeds for autonomous systems. MIT's RaceCar \cite{karaman2017project}, and University of Pennsylvania's F1/10 \cite{F1/10} have become popular autonomous racing platforms built on Traxxas 1/10 scale remote-controlled car with an NVIDIA's Jetson TX1 on-board computational unit. These cars use cameras, IMUs, and expensive LIDAR (\$1,775) systems for performing simultaneous localization and mapping (SLAM), whereas DeepNNCar performs e2e learning using data from a limited array of sensors (camera, IR-Optocoupler, LIDAR (\$100)). The cost comparison for F1/10 and DeepNNCar platforms is  \$3000 vs. \$518.

% Also, these cars are far more expensive compared to the cost of DeepNNCar (approx. \$3000 vs. \$518).

DeepPicar \cite{bechtel2017deeppicar} is another remote-controlled car platform that uses a smaller 1/24 scale chassis. Like DeepNNCar, this platform also uses a Raspberry Pi3 (RPi3) as the computational unit and performs e2e learning based autonomous driving using NVIDIA's DAVE-II CNN. This platform is relatively inexpensive (\$70) but has a considerably smaller chassis and uses discrete steering actuation unlike DeepNNCar which performs continuous steering.  
Donkey car \cite{roscoe3donkey} is an open source autonomous car platform for small-scale remote-controlled cars. These cars are built on a 1/16 or 1/10 scale chassis, and use RPi3 as the computational unit along with a wide angle RPi camera that is the primary sensor. Like the other e2e learning platforms these cars use a 7-layer CNN, which take in image inputs and predict categorical throttle values (i.e. discreet steer and speed actuation controls). In comparison to our platform, this car uses a different CNN model, and it performs discreet control actuations.

Compared to the existing platforms, DeepNNCar has an ideal tradeoff of cost vs. autonomous driving functionalities.

\section{DeepNNCar: Testbed for Autonomous Driving}
\label{sec:deepnncar}

\begin{figure}[t]
\centering
 \includegraphics[width=0.9\columnwidth]{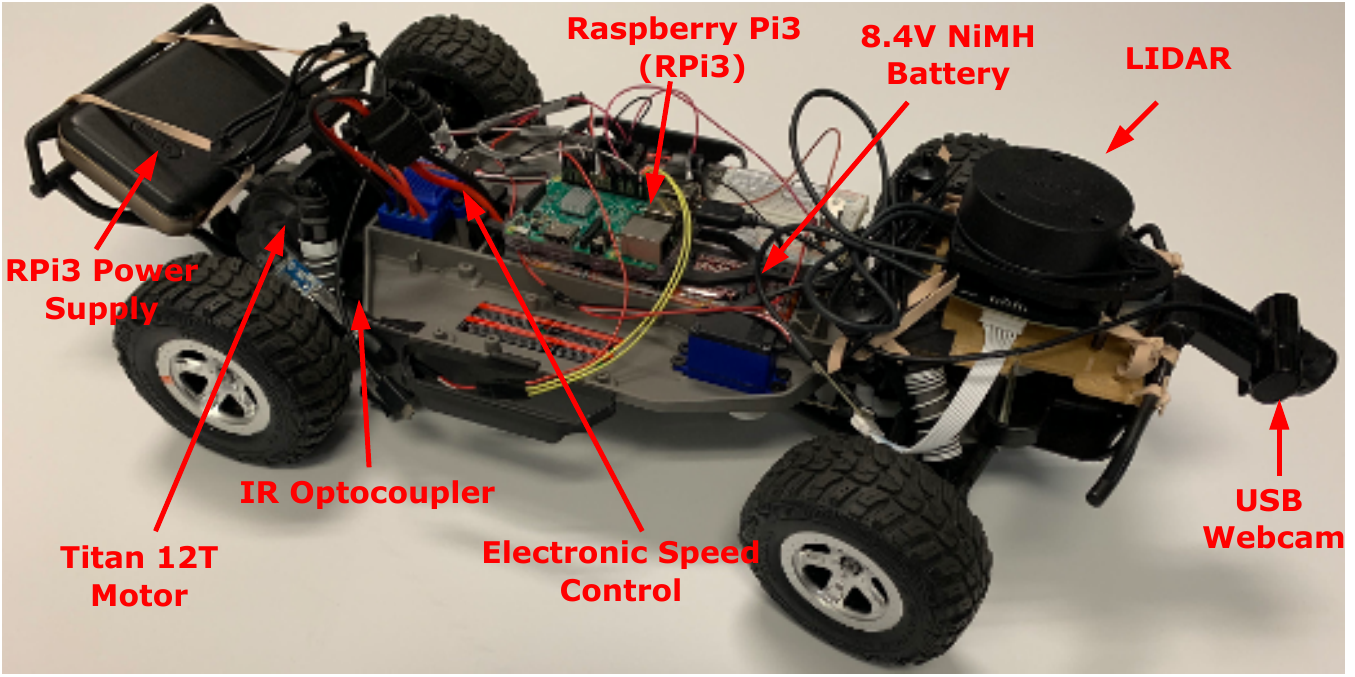}
 \caption{DeepNNCar is a resource constrained remote-controlled car that is designed to perform end-to-end learning based autonomous driving. The hardware components and operating modes of the car is discussed in \cref{sec:deepnncar}.}
 \label{fig:DeepNNCar}
 \vspace{-0.3em}
 \end{figure}

To understand the methodology discussed in later sections, we first introduce DeepNNCar\footnote{\label{github}Build instructions, source code, and videos of DeepNNCar can be found at: \url{https://github.com/scope-lab-vu/deep-nn-car}} (in \cref{fig:DeepNNCar}) that is built using the chassis of the Traxxas Slash 2WD 1/10 scale remote-controlled car. It has two on-board motors - a servomotor for steering control, and a Titan 12T 550 motor for motive force - both of which are powered by an 8.4 V NiMH battery. A Raspberry Pi 3 (RPi3) is the onboard computational unit which performs all the required computations and interfaces with the sensors. RPi3 reserves two GPIO pins to generate Pulse Width Modulation (PWM) signals that are used to control the motors of the car. The PWM signal is defined by a duty cycle component and a frequency component. The duty cycle of a PWM signal is the proportion of time the signal remains in the high state (or logical 1) over the total time it takes to complete one cycle. 

For the DeepNNCar the duty cycle is the percentage of a digital square pulse with a period 10 ms (frequency = 100 Hz). Varying the PWM duty cycle percentage allows us to control the two onboard motors of the car. For the servomotor, varying the duty cycle $\in$ [10\%, 20\%] results in a continuous steering angle $\in$ [-30\textdegree, 30\textdegree]. Similarly, for the titan motor, varying the duty cycle $\in$ [15.58\%, 15.70\%] results in a speed $\in$ [0, 1] m/s. Throughout this work we use the notations ($\nu$) for speed and ($\theta$) for steering. The speed is always refereed in-terms of m/s, and steering in degrees. However, to control the testbed, the steering and speed are varied as PWM duty cycle values. 

The goals of DeepNNCar are: (1) minimize the soft constraint violations, i.e. reduce the out-of-track occurrences, and (2) optimize the speed based on the different track segments. i.e. each track segment has different maximum attainable speed (straight segment -- 0.65 m/s, curved segment -- 0.25 m/s), and the car must learn to switch between the speed modes.

\begin{figure*}[t]
\centering
 \includegraphics[width=\textwidth]{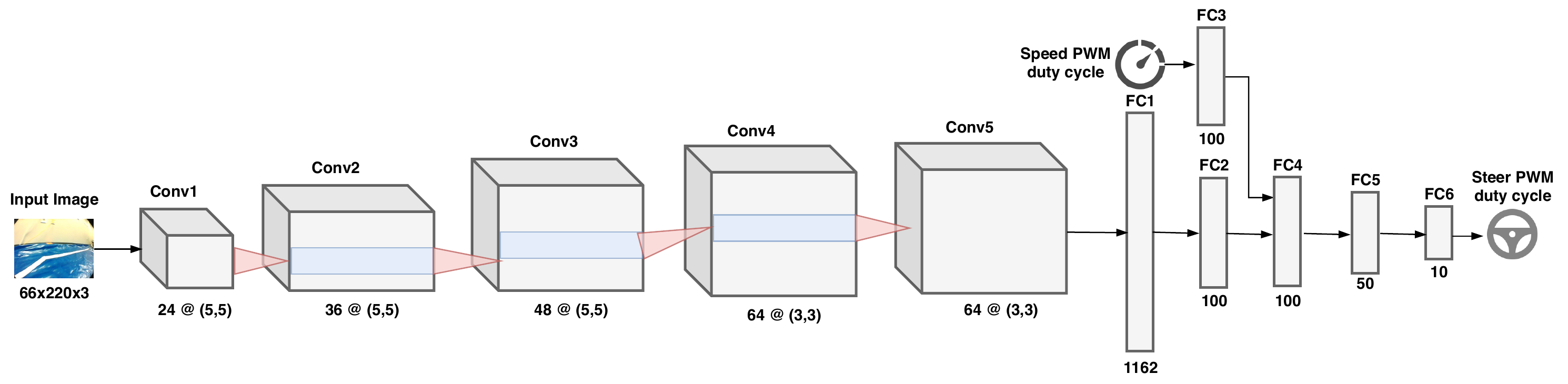}
 \caption{Modified NVIDIA's DAVE-II CNN, which takes in images and speed PWM duty cycle as input and predicts the steering PWM duty cycle. This modified model also takes in the speed as the inputs, which is not considered by the original NVIDIA DAVE-II model \cite{bojarski2016end}. The model has five convolutional layers and six fully connected layers. Conv-- represents the convolutional layers and sizes of the filters are mentioned below. FC -- represents the fully connected layers and the number of neurons is mentioned below.}
 \label{fig:CNN}
 \vspace{-0.1in}
 \end{figure*}
 
%  \textcolor{red}{this figure should be included as a pdf and should be made sharpt. cite the dave-II paper here.}

\subsection{Sensors and Operating Modes}

\ul{Sensors}: A USB webcam is attached to the RPi3 to capture images at 30 Frames Per Second (FPS) with a resolution of 320 $\times$ 240 RGB pixels. A slot-type IR Opto-coupler is attached to the chassis near the rear wheel that counts the revolutions of the wheel. The speed of the car is calculated based on the frequency of revolutions. The captured RGB images and the speed values are recorded on the computational unit and later utilized during the data collection, and training modes. 

\ul{Operating Modes}: DeepNNCar (server) is used along with a desktop (client) to provide a server-client setup for data collection, monitoring and runtime diagnostics. An Xbox controller is connected to the desktop via Bluetooth and it communicates with the car using TCP messages. Using this setup there are three different modes in which the car can function: (1) data collection -- manually driving the car to collect training data (images, steering PWM duty cycle, and speed PWM duty cycle), (2) autonomous driving -- for autonomous driving using LEC or other simplex strategies discussed in this work, and (3) livestream tracking -- for streaming runtime images captured by the camera to the client for runtime tracking.   

\subsection{Vehicle Control}
Two independent controllers were developed for the DeepNNCar: one using a \ac{lec} and the other using traditional image processing algorithms provided by OpenCV. Both controllers took the available camera images as input and were required to output PWM duty cycle values for both steering and speed control. 

The speed control while driving the DeepNNCar using the LEC and OpenCV controller is initially set by a human supervisor. It is then controlled either using a constant throttle or a PID controller. Each of these controllers is discussed in more detail in the following sections.

\begin{figure}[t]
\centering
\includegraphics[width=0.95\columnwidth]{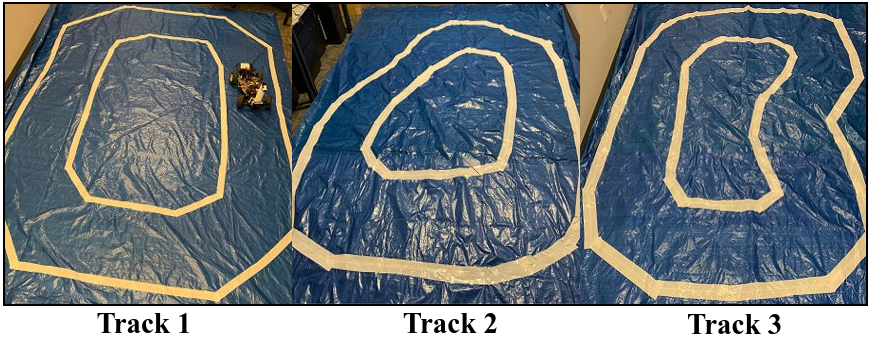}
\setlength{\belowcaptionskip}{-10pt}
\caption{Track 1 and Track 2: on which the training data for the LEC was collected. The RL exploration for the dynamic-weighted simplex strategy was performed on Track 1; Track 3: used to test the performance and accuracy of the trained LEC, and the  dynamic-weighted simplex strategy. The tracks were built indoor in our laboratory using 10' x 12' blue tarps. The videos of DeepNNCar preforming on these three tracks can be found in \protect\url{https://github.com/scope-lab-vu/deep-nn-car}}.
\label{fig:ClientFeedback}
% \vspace{-0.4em}
\end{figure} 

\subsubsection{LEC Controller}
In the current implementation, the hardware performs autonomous driving based on \ac{e2e} learning. For this we use a modified version of NVIDIA's DAVE-II CNN that takes in image and PWM duty cycle of speed $\nu$ as inputs to predict the PWM duty cycle of steering ($\theta_L$). This is an extension to the original DAVE-II CNN \cite{bojarski2016end} that took in only image as input to predict steering $\theta_L$.

\ul{Network Architecture}: Our CNN model (in \cref{fig:CNN}) has five convolutional layers and six fully connected layers. We modified the CNN architecture for two reasons. First, the steering and speed actions cannot be treated independently. Any change in the speed will impact the steering performance. Through experiments we observed that the modified CNN takes wider trajectories at turns compared to the original one. Second, since the quality of the captured image deteriorates as speed increases, additional information is required for the CNN to predict correct steering values. 

\ul{Model Training and Validation}: We train the CNN with 6000 labeled images collected from Track 1 and Track 2 (in \cref{fig:ClientFeedback}). We also performed hyperparameter tuning \cite{bergstra2011algorithms} using random search \cite{bergstra2012random} to select the appropriate learning rate, and number of epochs.

We validated the trained model using a test dataset of 1000 images. We then used Mean Square Error (MSE) \cite{willmott2005advantages} as a metric to quantify the distance between the actual and predicted steering values. Once the MSE was within our acceptable limits (0.1 -- found through varied experiments), we deploy the model on the car for autonomous driving.

\subsubsection{OpenCV Controller}
The OpenCV controller is designed using classical image processing algorithms and it performs two tasks: (1) lane detection -- finds lanes in images using classical lane detection algorithm, and (2) steer computation -- associates a discreet steering ($\theta_C$) based on the lanes detected. The output of these tasks are the detected lanes, the lane segments ($\hat M$) identified from number of lanes, the discreet steering $\theta_C$ (in degrees) and its corresponding PWM duty cycle, and a stop signal alarm when the car runs out of the track. 

\ul{Lane Detection (LD) algorithm:} We convert the 200 $\times$ 66 RGB image to gray scale (to reduce computation time) and then apply the LD algorithm. The different image processing involved in the LD process are: 

\begin{itemize}[leftmargin=*,noitemsep]
    \item \underline{Gaussian blur and white masking}: A 3x3 Gaussian kernel is convolved across the image to reduce noise. Next, all pixels except those within a specified range (e.g., [215, 255]) are masked, thus differentiating the track lanes from the foreground.
    \item \underline{Canny edge detection \cite{canny1986computational}}: The algorithm first computes a gradient of pixel intensities. An upper and lower threshold of these gradients is defined at compile time. A comparison of the pixel gradients to these thresholds in addition to hysteresis (suppress all weak and unconnected edges) can determine if a pixel is an edge or not. The edges reveal the boundary of the lanes.
    \item \underline{Region of interest (ROI) selection}: The image is divided into two similar 30x66 regions of interest to capture the left and right lane respectively. 
    \item \underline{Hough line transform \cite{illingworth1988survey}}: A Hough line transform is applied to each ROI to detect the existence of a lane based on the results of the canny edge detection algorithm. Using this information, we determine a label for the track segment.
\end{itemize}

\begin{figure}[t]
\centering
\includegraphics[width=\columnwidth]{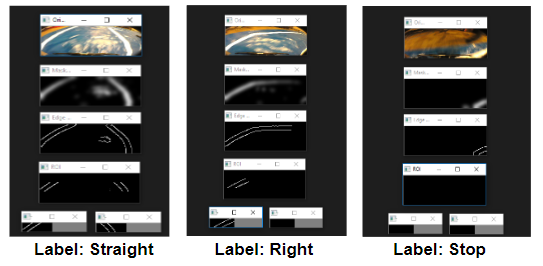}
\setlength{\belowcaptionskip}{-10pt}
\setlength{\abovecaptionskip}{-4pt}
\caption{Image transformations as it passes through the different steps of the LD algorithm. Based on the lanes detected the OpenCV controller assigns a track segment and then associates a discreet steering value. During the livestream mode, real-time images captured from the car are relayed to the client, which runs LD algorithm continuously to get this visualization.}
\label{fig:lanedetection}
\end{figure}

\cref{fig:lanedetection} shows the image transformation as it passes through the different steps of the LD algorithm. As seen from the figure, when the OpenCV controller detects two lanes it classifies the segment to be straight. Similarly, if it finds only the left lane, it classifies the segment to be right. Finally, if it does not find both the lanes, it classifies the track segment to be out and issues a stop signal.

Once the lanes are detected, the OpenCV controller associates a discrete steering angle $\theta_{C}$, and its corresponding PWM duty cycle (similar to  \cite{mcfall2016using}): if two lanes are detected, $\theta_{C}= 0^{\circ}$ (corresponds to a PWM duty cycle of 15\%); if the right lane is detected, $\theta_{C}=-30^{\circ}$ (corresponds to a PWM duty cycle of 10\%); and if the left lane is detected, $\theta_{C}=30^{\circ}$ (corresponds to a PWM duty cycle of 20\%).

% The precision-recall graph in \cref{fig:PR} represents the

The LD algorithm was tested with a dataset of 3000 images and correctly labeled the track regions with an accuracy of 89.6\%. 
\cref{Table:PR} summarizes the accuracy of the LD algorithm in correctly classifying the lanes in to straight or curved (left and right) segments. To generate this plot, we manually iterated through the 3000 images and made a note of the actual lane segment visualized by a human supervisor, and the lane segment as classified by the LD algorithm. Using this we gathered the number of correct lane predictions and mis-classifications to generate the data for the precision-recall graph. Using the precision and recall values we also computed the F1 score \cite{goutte2005probabilistic} - a metric which indicates the accuracy of the LD algorithm. The F1 scores for the straight and curved segments were 92.4\% and 91.8\% respectively. Higher F1 score indicates the LD algorithm's accuracy in classifying the lane segments correctly.  

\begin{table}[h!]
\centering
\footnotesize
\begin{tabular}{|c|c|c|}
\hline
Track Segment & Precision (\%) & Recall (\%) \\ \hline
Straight      & 97.73         & 87.78      \\ \hline
Curved        & 90.78         & 93.05      \\ \hline
\end{tabular}%
\caption{The precision and recall values to evaluate the performance of the LD algorithm in different segments of the track. We manually iterated through 3000 images and compared the actual lane segments to those predicted by the LD algorithm.}
    \label{Table:PR}
\end{table}

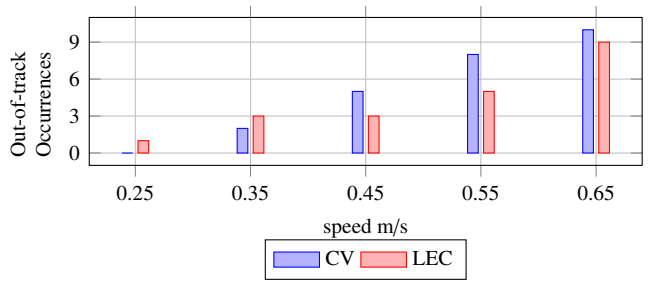
\begin{figure}[t]
% \vspace{-0.1in}
    \centering
    \setlength{\belowcaptionskip}{-6pt}
\begin{tikzpicture}
\begin{axis}[
  ybar,
  area legend,
    font=\footnotesize,
  ytick={0,3,6,9,12,15},
  bar width=4pt,
  width=\columnwidth,
  grid=major,
  height=0.4\columnwidth,
  ylabel style={text width=1.75cm}, ylabel=Out-of-track\\ Occurrences\\,
  xlabel=speed m/s,
  legend style={at={(0.5,-0.5)},anchor=north,legend columns=-1},
  symbolic x coords=
    {0.25, 0.35, 0.45, 0.55, 0.65},
  xtick=data,
  ]
  \addplot coordinates {(0.25,0)(0.35,2)(0.45,5)(0.55,8)(0.65,10)};
  \addlegendentry{CV};
     \addplot coordinates {(0.25,1)(0.35,3)(0.45,3)(0.55,5)(0.65,9)};
     \addlegendentry{LEC};
    %   \addplot coordinates {(0.25,0)(0.35,0)(0.45,1)(0.55,3)(0.65,7)};
    %   \addlegendentry{FW-WSA};
    %      \addplot coordinates {(0.25,0)(0.35,0)(0.45,0)(0.55,1)(0.65,3)};
    %      \addlegendentry{DW-WSA};
\end{axis}
\end{tikzpicture}
\setlength{\belowcaptionskip}{-10pt}
  \caption{The out-of-track occurrences for different speeds in the curved segments of the track. From figure, CV: driving only with the OpenCV controller, LEC: driving only with the modified Dave-II model. The horizontal axis shows the different speeds of the car during the experiment. The data is collected by running DeepNNCar with each of these controllers independently around the track for 10 laps. The out-of-track occurrences was manually noted down by a human supervisor.}
    \label{fig:lecperformance}
    % \vspace{-0.1in}
\end{figure}

\subsection{Performance of the Controllers}
\label{sec:performance}
We evaluate the performance of these controllers based on their highest achievable speeds and the number of out-of-track occurrences. During the testing regiment, the DeepNNCar drives 10 laps each around the track for both the trained LEC controller and the OpenCV controller. The track was divided into straight and curved (left and right) segments, and performance results were compiled for each segment. The out-of-track occurrences was manually noted by a human supervisor. In the straight segment the LEC controller performed well with very few (typically 2 or 3) out-of-track occurrences up to a speed of 0.55 m/s. The OpenCV controller had typically 1 or no out-of-track occurrences up to 0.35 m/s. Above 0.35 m/s the OpenCV controller had higher chances of leading the car out-of-track as compared to the LEC controller. 

The out-of-track occurrences of the different controllers in the curved segment of the track is shown in \cref{fig:lecperformance}. The LEC has very few (typically 3) out-of-track occurrences up to a speed of 0.45 m/s and gradually as the speed increases the out-of-track occurrences increases. In comparison, the OpenCV controller performed well up to a speed of 0.35 m/s with only 1 out-of-track occurrences. Again, above 0.35 m/s the OpenCV controller started making higher wrong predictions leading the car out-of-track. At 0.65 m/s, the OpenCV controller had a higher out-of-track occurrence of 10 as compared to 8 of the LEC controller.

These results show that each controller outperforms the other controller in certain segments of operation. This implies that the total number of out-of-track occurrences can be reduced by intelligently blending the two controller actions. In the next section, we introduce the weighted simplex strategy for this purpose.

\section{Dynamic-Weighted Simplex Strategy}
\label{sec:Simplex Strategies}
Earlier, we discussed the two key challenges of using the Simplex Architecture, they are: (1) designing an effective decision logic, and (2) instantaneous switching between the controllers which may cause undesirable transient effects on system performance. In addition, the OpenCV controller did not perform reliably at high speeds (> 0.35 m/s) in the curved segments of the track (discussed in \cref{sec:performance}). So, the OpenCV controller cannot always be relied on as a safe controller when the vehicle is operating at high-speed in curved track segments.

To overcome these challenges, we compute the systems output as a weighted blending of the two controller outputs. This means the weights in \cref{Eqn:gegeralsimplex} are no longer restricted to taking only values of 0 or 1, but can take any value in the continuous range [0,1]. However, the combination of the ensemble weights must sum to 1. Such blending approaches have been popularly used in model ensembles \cite{jimenez1998dynamically}, and multiple model adaptive predictive control \cite{zhang2012stable}. We call this approach the ``weighted simplex strategy''. 

As shown in \cref{fig:wss}, the weighted simplex strategy works like the conventional Simplex Architecture in the high performance and the safe regions. The targeted spectrum of this strategy is the arbitration region, which is represented by the yellow grid in \cref{fig:wss}. In this region, the conventional Simplex Architecture performs an instantaneous transition from the AC to the BC. In contrast, the continuous ensemble weights of the weighted simplex strategy can be used to provide a smoother transition between the two controllers. However, one key challenge in using this strategy is the calculation of appropriate ensemble weights.  

% Still, calculating appropriate ensemble weights is one key challenge of this strategy.

% In the transition region (see \cref{fig:wss}), using dynamic weights would provide continuous transition, whi  

For this, we present a strategy called the ``dynamic-weighted simplex strategy" which computes dynamic ensemble weights. This strategy uses an RL technique to find the optimal dynamic ensemble weights for the different segments of the track. In this work, we demonstrate the RL based ensemble weight selection process for DeepNNCar. The car's steering is computed as the weighted ensemble of the steering values computed by the LEC controller and the OpenCV controller. The weighted simplex combination for DeepNNCar steering is computed using \cref{Eqn:DeepNNCar}. 

\begin{equation}
 \theta_{D} = W_L \cdot \theta_{L} + W_C \cdot \theta_{C}
 \label{Eqn:DeepNNCar}
 \end{equation}
 
Where, $\theta_{D}$ represents the PWM duty cycle corresponding to steering computed while using dynamic weights, $W_L$ is the ensemble weight to the LEC, $W_C$ is the ensemble weight to the OpenCV controller, $\theta_L$ is the PWM duty cycle corresponding to steering computed by the LEC controller and $\theta_C$ is the PWM duty cycle corresponding to steering computed by the OpenCV controller.

% Where, $\theta_{D}$ represents the steering PWM duty cycle while using dynamic weights, $W_L$ is the ensemble weight to the LEC, $W_C$ is the ensemble weight to the OpenCV controller, $\theta_L$ is the steering PWM duty cycle computed by the LEC controller and $\theta_C$ is the steering PWM duty cycle computed by the OpenCV controller.

Other than selecting the $W_L$ and $W_C$, the reward function (explained in detail in \cref{sec:learn}) in the RL setup is also designed to control the speed of the car. The PWM duty cycle of the new speed ($\nu_{t+1}$) can be incremented or decremented by ($\delta \nu$) based on the PWM duty cycle of the system's current speed ($\nu_{t}$).

\begin{equation}
 \nu(t+1) = \nu_{(t)} \pm \delta \nu
 \label{Eqn:speedvariation}
 \end{equation}
 
\subsection{The Learning Approach}
\label{sec:learn}
The key contribution of our work is to find dynamic weights that optimally balances the performance and safety across all the segments of the track. As discussed earlier, we use RL to compute the dynamic ensemble weights. The continuous learning based setup of RL encouraged us to use it for our application, which has a dynamically changing environment (changing track segments). Any RL algorithm can be used in the decision logic, however in this work, we use the simplest of RL algorithms called the Q-learning \cite{watkins1992q} algorithm. Through this section we discuss the important elements and steps required to setup the Q-learning problem. The important elements required to setup the Q-learning problem are:

\ul{Environment}: estimate the state (s) of a system based on an internal estimate of the Markov Decision Process (MDP) and computes a reward value (r) for each action (a) produced by an RL component. In the DeepNNCar, the environment is implemented as a standalone component within our middleware framework (described in the next section).

\ul{State (s)}: represents the current state of the system. For the DeepNNCar, the system state is continuously changing as the car interacts with the environment (track). Since our problem is to find optimal ensemble weights and optimal speed, we encode weights ($W_L$, $W_{C}$), the PWM duty cycle corresponding to speed ($\nu$), and the PWM duty cycle corresponding to steering $\theta_L$ and $\theta_C$ as the state information. As we do not have an explicit sensor for locating the position of the car, we use the steering values $\theta_L$ and $\theta_C$ to identify the position, and thus it is included as internal state information. The state $s_t$ of the car at time t is: ($W_{L(t)}$, $W_{C(t)}$, $\nu_{(t)}$, $\theta_L$, $\theta_{C}$). The transition between the states happen when the element $W_L$ changes by $\pm$ $\delta W_L$, $W_{C}$ changes by $\pm$ $\delta W_C$, and the corresponding PWM duty cycle of $\nu$ changes by $\pm$ $\delta \nu$. These transitions are illustrated in \cref{eqn:transition}. 

% A sample possible transition in the action space is illustrated in \cref{Table:actionspace}.

% \begin{figure}[t]
% \setlength{\belowcaptionskip}{-6pt}
%     \centering
%     \includegraphics[width=0.8\columnwidth]{figures/transition-diagram4.pdf}
%     \caption{The elements of the state are shown. The transition between the states happen when the elements $\delta W_L$, $\delta W_{C}$ change by $\pm$ 0.05, and the corresponding PWM duty cycle of $\delta \nu$ changes by $\pm$ 0.001. A small part of the state transitions is shown in \cref{Table:actionspace} \textcolor{red}{this is too simple of a figure - explain this figure as an equation rather than figure. }}
%     % \vspace{-0.4mm}
%     \label{fig:transition}
% \end{figure}

\begin{equation}
\centering
\begin{bmatrix}
    W_{L(t+1)}\\
    W_{C(t+1)}\\
    \nu_{(t+1)}
\end{bmatrix}
= 
\begin{bmatrix}
    W_{L(t)} \\
    W_{C(t)}\\
    \nu_{(t)}     
\end{bmatrix} 
\pm
\begin{bmatrix}
    \delta W_L \\
    \delta W_C \\
    \delta \nu     
\end{bmatrix} 
\label{eqn:transition}
\end{equation}

Further, to limit the dimension of our state space we reduce the number of states in our MDP, by discretizing the elements of the state. Each weight $W_L, W_{C}\in[0,1]$ is divided into 21 elements ($W_C$ = 1-$W_L$, as the two weights sum up to 1) in increments of 0.05. Also, our tracks were small and we could not maintain the car within the track at speed $>$ 0.65 m/s. Thus, we restricted the speed $\in$ [0, 0.65] m/s, and the corresponding PWM duty cycle to $\nu \in$ [15.58\%,15.62\%]. We then divided this PWM range into 41 elements in increments of 0.001. 

\begin{table*}[t]
\resizebox{\textwidth}{!}{%
\begin{tabular}{|l|l|l|l|l|l|l|l|l|l|l|l|}
\hline
                                                               & \begin{tabular}[c]{@{}l@{}}$\theta_{L}$= 10\%\\    (Left)\end{tabular} & $\theta_{L}$= 11\% & $\theta_{L}$= 12\% & $\theta_{L}$= 13\%                                                  & $\theta_{L}$= 14\%                                                  & \begin{tabular}[c]{@{}l@{}}$\theta_{L}$= 15\%\\  (Straight)\end{tabular} & $\theta_{L}$= 16\%                                                  & $\theta_{L}$= 17\%                                                  & $\theta_{L}$= 18\%                                                  & $\theta_{L}$= 19\%                                                  & \begin{tabular}[c]{@{}l@{}}$\theta_{L}$=20\%\\   (Right)\end{tabular} \\
\hline                                                              
\begin{tabular}[c]{@{}l@{}}$\theta_{C}$ = 20\%\\    (Right)\end{tabular} & -                                                            & -        & -        & -                                                         & \begin{tabular}[c]{@{}l@{}}$W_L$=0.95\\ $W_C$=0.05\end{tabular} & \begin{tabular}[c]{@{}l@{}}$W_L$=0.95\\ $W_C$=0.05\end{tabular}      & \begin{tabular}[c]{@{}l@{}}$W_L$=0.95\\ $W_C$=0.05\end{tabular} & \begin{tabular}[c]{@{}l@{}}$W_L$=0.85\\ $W_C$=0.15\end{tabular} & \begin{tabular}[c]{@{}l@{}}$W_L$=0.80\\ $W_C$=0.20\end{tabular} & \begin{tabular}[c]{@{}l@{}}$W_L$=0.80\\ $W_C$=0.20\end{tabular} & \begin{tabular}[c]{@{}l@{}}$W_L$=0.80\\ $W_C$=0.20\end{tabular}   \\
\hline
\begin{tabular}[c]{@{}l@{}}$\theta_{C}$= 15\%\\  (Straight)\end{tabular} & -                                                            & -        & -        & -                                                         & \begin{tabular}[c]{@{}l@{}}$W_L$=0.95\\ $W_C$=0.05\end{tabular} & \begin{tabular}[c]{@{}l@{}}$W_L$=0.95\\ $W_C$=0.05\end{tabular}      & \begin{tabular}[c]{@{}l@{}}$W_L$=0.90\\ $W_C$=0.10\end{tabular} & -                                                         & -                                                         & -                                                         & -                                                            \\
\hline
\begin{tabular}[c]{@{}l@{}}$\theta_{C}$= 10\%\\    (Left)\end{tabular}   & -                                                            & -        & -        & \begin{tabular}[c]{@{}l@{}}$W_L$=0.80\\ $W_C$=0.20\end{tabular} & -                                                         & -                                                              & -  
& \begin{tabular}[c]{@{}l@{}}$W_L$=0.90\\ $W_C$=0.10\end{tabular} & -                                                         & -                                                         & - \\\hline 
\end{tabular}
}
\caption{The variations in the ensemble weights are captured with respect to the change in the steering PWM duty cycle of LEC and OpenCV controller. The rows indicate the discretized steering PWM duty cycle of the OpenCV controller $\theta_C$ $\in$ [10\%,15\%,30\%]. The columns indicate the steering PWM duty cycle of the LEC controller $\theta_L$ $\in$ [10\%,20\%]. These steering values are discretized in steps of 1\%. The blocks with ``-" indicate those steering value combinations were not encountered in our experiments. The prime reason for this was our track was small and did not have very steep left turns. A large track with lots of turns could have better explored all the combinations. Also, it is evident from top right corner of the table that, $W_{C}$ starts to increase as car starts turning in the right segment.}
\label{Table:varyweights}
\end{table*}

% \textcolor{red}{use different shades of gray. Write a - in places that are empty. Explain why the space is no unexplored. why are there gaps. and what will happen if the car operates in those areas. construct this figure as a table using latex table generator}

\ul{Reward}: Designing an appropriate reward function is the key step in the Q-learning process. Our goal is to reduce the soft constraint violations while improving the performance of the system. Thus, we need to incorporate them in designing the reward function. Our reward function is a combination of the performance factor and the safety factor. In the case of the DeepNNCar, we chose speed to be the performance factor and a deviation of the car from the center of the track ($\hat t$) to be the safety factor. The calculated reward is expressed in \cref{eqn:reward}:

\begin{equation}
r(s_t,a_t) = \nu_t \cdot (1 - \hat t) 
\label{eqn:reward}
\end{equation}

where, $\nu_{t}$ is the current speed of the car and $\hat t$ is a scalar quantity calculated based on the deviation of the car from the center of the track. The measure $\hat t$ is calculated based on the lane segment information given by the LD algorithm. If the algorithm detects both lanes in the captured image, we infer that the car is at the center of the track and we assign $\hat t = 0$. If the algorithm detects only one lane, we assume the car has deviated from the center and we assign $\hat t = 1/2$. Finally, if no lanes are detected, we assume the car has moved out of the track and we assign $\hat t = 10$, a large penalty.

The positive reward given for remaining in the center of the track encourages the car to always select an action that keeps it within the track boundaries. Also, the reward value increases proportionately to vehicle speed which encourages the RL component to optimize for the highest achievable speed without exiting the track boundaries.

% \begin{table}[h!]
% \setlength{\tabcolsep}{5pt}
% \centering
%  \footnotesize
%     \setlength{\belowcaptionskip}{-2pt}
%     \begin{tabular}{|p{1.25cm}|p{2.05cm}|p{2.05cm}|p{2.05cm}|}
%     \hline
%     action\newline space &$\uparrow \nu_t\ by\ 0.001$ & $\downarrow \nu_t\ by\ 0.001$ & NOP\\
%     \hline
%     $\uparrow W_L\ by\ 0.05$ & (0.95,0.05,15.591,\newline 16,15) & (0.95,0.05,15.589,\newline 16,15) & (0.95,0.05,15.590\newline 16,15) \\
%     \hline
%     $\downarrow W_L\ by\ 0.05 $ & (0.85,0.15,15.591\newline 16,15) & (0.85,0.15,15.589\newline 16,15) & (0.85,0.15,15.590\newline 16,15)\\
%     \hline
%     NOP & (0.90,0.10,15.591\newline 16,15) &(0.90,0.10,15.589\newline 16,15) &(0.90,0.10,15.590\newline 16,15)\\
%     \hline
%     \end{tabular}
%     \caption{Action space for a given state ($W_L=0.90$, $W_{C}=0.10$, $\nu_t=15.590\%$, $\theta_L=16\%$, $\theta_C$=15\%). Similar action combinations are generated for other states. NOP: means no operation. This is a sample action space when the car is in the straight segment of the track. So, the steering values remain almost the same in the next achievable states too.}
%     \label{Table:actionspace}
%     \vspace{-0.05in}
% \end{table}

% \usepackage{graphicx}
\begin{table}[h!]
\centering
 \footnotesize
\begin{tabular}{|c|c|c|c|}
\hline
\begin{tabular}[c]{@{}c@{}}Action\\ Space\end{tabular}        & $\uparrow$ $\nu_t$ by 0.001                                                         & $\downarrow$ $\nu_t$ by 0.001                                                         & NOP                                                                 \\ \hline
\begin{tabular}[c]{@{}c@{}}$\uparrow$ $W_L$ \\ by 0.05\end{tabular} & \begin{tabular}[c]{@{}c@{}}(0.95,0.05,15.591,\\ 16,15)\end{tabular} & \begin{tabular}[c]{@{}c@{}}(0.95,0.05,15.589,\\ 16,15)\end{tabular} & \begin{tabular}[c]{@{}c@{}}(0.95,0.05,15.590,\\ 16,15)\end{tabular} \\ \hline
\begin{tabular}[c]{@{}c@{}}$\downarrow$ $W_L$ \\ by 0.05\end{tabular} & \begin{tabular}[c]{@{}c@{}}(0.85,0.15,15.591,\\ 16,15)\end{tabular} & \begin{tabular}[c]{@{}c@{}}(0.85,0.15,15.589,\\ 16,15)\end{tabular} & \begin{tabular}[c]{@{}c@{}}(0.85,0.15,15.590,\\ 16,15)\end{tabular} \\ \hline
NOP                                                           & \begin{tabular}[c]{@{}c@{}}(0.90,0.10,15.591,\\ 16,15)\end{tabular} & \begin{tabular}[c]{@{}c@{}}(0.90,0.10,15.589,\\ 16,15)\end{tabular} & \begin{tabular}[c]{@{}c@{}}(0.90,0.10,15.590,\\ 16,15)\end{tabular} \\ \hline
\end{tabular}%
\caption{Action space for a given state ($W_L=0.90$, $W_{C}=0.10$, $\nu_t=15.590\%$, $\theta_L=16\%$, $\theta_C$=15\%). $W_C$ is computed as (1-$W_L$). Similar action combinations are generated for other states. NOP: means no operation. This is a sample action space when the car is in the straight segment of the track. So, the steering values remain almost the same in the next achievable states too.}
    \label{Table:actionspace}
\end{table}

\ul{Agent}:
%Is the \ac{rl}-based component which must learn to select an optimal action based on the state information provided by the environment.
In our setup the DeepNNCar is the agent. For each state $s\in S$, the agent performs an action $a \in A$, which results in a reward, $r: S\times A \rightarrow \mathbb{R}$, as the agent transitions from the initial state to a new state $s \rightarrow s'\in S$. An action space is created for all the different combinations of the state. As an example, the possible action space for the DeepNNCar when starting from the state $s\ =\ $ ($W_L=0.90$, $W_{C}=0.10$, $\nu_t=15.590\%$, $\theta_L=16\%$, $\theta_C$=15\%) is shown in \cref{Table:actionspace}. Since, there are three possible actions $\delta W_L$, $\delta W_C$ and $\delta \nu$, there are 9 possible actions that can be performed from any state.

\subsection{Exploration}
Exploration is the training phase of RL, where the RL agent learns to select an appropriate action by continuously interacting with the environment. During this phase, the car stores results as state-action pairs $Q(s_t,a_t)$, which is a measurement of the quality of the immediate reward $r(s_t,a_t)$ from being in state $s_t$ and taking action $a_t$ discounted by the maximum expected future award in the new state denoted $Q'(s_{t+1},a_{t+1})$. The new Q-state, $Q'(s_{t+1},a_{t+1})$, is obtained by selecting an action which results in the maximum Q-value as shown in the \cref{eqn:newstate} below.

\begin{equation}
    Q'(s_{t+1},a_{t+1}) = \max_{a_k\in A}Q(s_{t+1},a_k)
    \label{eqn:newstate}
\end{equation}

The Q-state is updated using the Bellman equation which takes the current state and action as inputs along with the parameters $\alpha \in[0,1]$ and $\gamma \in[0,1]$. $\alpha$ controls the learning rate of the algorithm, while $\gamma$ represents the discount factor which balances the importance of future benefits over immediate benefits (i.e. decreasing $\gamma$ increases priority on obtaining immediate rewards).  

\small
\begin{multline}
  Q_{new}(s_t,a_t) = Q(s_t,a_t) + \alpha[r(s_t,a_t) + \\\gamma \cdot \max{Q'(s_{t+1},a_{t+1})} - Q(s_t,a_t)]
\end{multline}
\normalsize
    
The new Q-states and the Q-value calculated from the Q-learning algorithm are stored in a Q-Table, which is a look up table that holds the state-action pairs $Q(s_t,a_t)$ and the associated reward $r(s_t,a_t)$. For our experiments we used a learning rate $\alpha = 0.1$, discount factor $\gamma = 0.4$, and 1000 exploration steps. These hyper-parameters were obtained by tuning through various training runs. Also, for all the exploration runs we fix $W_L$=0.5, and $W_{C}$=0.5 as the initial state, and we start from different parts of the tracks. We hypothesize doing this could help us obtain a robust Q-Table with larger percentage of the state-action pairs explored.

% as the state information is encoded in terms of weights, during exploration we start with $W_L$=0.5, and $W_{C}$=0.5 as the initial states.

% \begin{figure}[t]
% \setlength{\belowcaptionskip}{-6pt}
% \setlength{\belowcaptionskip}{-4pt}
%     \centering
%     \includegraphics[width=\columnwidth]{figures/varying-weights3.pdf}
%     \caption{The variations in the ensemble weights are captured with respect to the change in the steering PWM duty cycle of LEC and OpenCV controller. Gray blocks -- when $\theta_{C}$=20\% (turning right), and $\theta_L$ $\in$ [10\%,20\%]. Green blocks -- when $\theta_{C}$=15 (driving straight), and $\theta_L$ $\in$ [10\%,20\%]. Yellow blocks -- when $\theta_{C}$=10\% (turning left), and $\theta_L$ $\in$ [10\%,20\%]. Most of the yellow blocks are empty as our tracks did not have steep left segments. It is evident from top right corner of the gray section that, $W_{C}$ starts to increase as car starts turning in the right segment. \textcolor{red}{use different shades of gray. Write a - in places that are empty. Explain why the space is no unexplored. why are there gaps. and what will happen if the car operates in those areas. construct this figure as a table using latex table generator}}
%     \vspace{-0.4mm}
%     \label{fig:varyweights}
% \end{figure}

\cref{Table:varyweights} shows the ensemble weights learned during exploration of the different steering values. In the table, the columns list the PWM duty cycle corresponding to the \ac{lec} controller steering, $\theta_L$ $\in$ [10\%,20\%], that is discretized in steps of 1. Similarly, the row lists the PWM duty cycle corresponding of the OpenCV controller steering, $\theta_C$ $\in$ [10\%,20\%], that has been discretized in steps of 5. The resulting rectangular regions is shown in \cref{Table:varyweights}. The learned ensemble weights for each region of operation are listed within the corresponding block. A block with a ``-" indicates a valid system state which was not represented in the training dataset, and thus no ensemble weights were learned. In addition, most of the blocks for $\theta_C$=10 are not explored. The reason for this is our tracks were small and did not have steep left turns. A larger track with different turns would result in a larger exploration. Also, when the car is driving in a straight segment (row with $\theta_C$=15), a higher weight is given to the output of the LEC. However, the OpenCV controller starts to get higher weights (see right corner of the row with $\theta_C$=20) as the car moves towards the right curved segment.

\subsection{Exploitation}
During exploitation the car uses the learned action sequence stored in the Q-Table. The state-action pair with the highest Q-value is chosen as the action of the current state. During exploitation the learnt Q-Table is loaded onto the car and based on the steering PWM values, the agent starts moving towards the appropriate weights. In straight segments the weights quickly move to $W_L$=0.95, and $W_{C}$=0.05, and in curved segments the weights hover around $W_L$=0.8, and $W_{C}$=0.2. 

If the state-space is not completely explored during the exploration phase, then there is a possibility that the agent may encounter unexplored states during exploitation and perform an incorrect action. This problem was encountered during our exploitation, causing occasional out-of-track occurrences. Since our tracks are small, 5 trial runs of 1000 steps exploration were enough to overcome this problem. Additionally, we had a decaying learning rate as the exploration run progressed. For systems with larger state spaces we would recommend a higher number of episodes and exploration steps. 

% Also, if safety is a major concern, then the system can be stopped when such an unknown state is encountered.

To compare the performance of the dynamic weighted approach, we also introduce a fixed-weighted simplex strategy. The fixed weights version uses static weights for the different segments of the tracks.

\subsection{Fixed-Weighted Simplex Strategy}
The fixed-weighted simplex strategy is an empirical weight selection mechanism. It uses fixed weights across the different segments of track. The weights are found through a trial-and-error approach by a human supervisor. For the DeepNNCar running on Tack 1 (see \cref{fig:ClientFeedback}), we found the optimal weights to be $W_L$=0.8, $W_C$=0.2. The weighted simplex steering equation using these weights is shown below:

\begin{equation}
\theta_{F} = 0.8 \cdot \theta_{L} + 0.2 \cdot \theta_{C}
\label{Eqn:argue}
\end{equation}

where, $\theta_{F}$ represents the PWM duty cycle corresponding to steering computed while using fixed weights, $\theta_L$ is the PWM duty cycle corresponding to steering computed by the LEC controller and $\theta_C$ is the PWM duty cycle corresponding to steering computed by the OpenCV controller.

To implement the fixed weighted strategy on the DeepNNCar we use the concept introduced by Fridman et. al. \cite{fridman2017arguing} called ``Arguing Machines'' with a slight modification. Their work focused on combining an LEC based system with a human supervisor to aid its decision making during uncertain situations. Their approach had two LEC controllers predicting steering values, and when the difference (or argument) between their predicted steering values is higher than a predefined threshold ($\tau_{SW}$), then the steering action is performed by a human driver. We replicate the same idea for DeepNNCar, but the human driver is replaced by \cref{Eqn:argue}. In our case, if the computed steering difference between the LEC controller and the OpenCV controller is greater than $\tau_{SW}$, then \cref{Eqn:argue} is used to compute the steering of the car. However, if the difference is lower than $\tau_{SW}$, then the predicted steering of the LEC is chosen to drive the car.

Further, the speed of the car is also calculated based on the argument between the controllers. If the difference among the controller predictions is greater than $\tau_{SW}$, then $\nu_t$ is decremented using \cref{Eqn:speedvariation}. However, if the  difference among the controller predictions is lesser than $\tau_{SW}$, then $\nu_t$ is incremented as it suggests consensus among the controllers predictions.

\subsection{Discussion}
It is evident from the above sections that computing weights dynamically based on the operating contexts (like changing track segments) is better than the fixed weights. The main reason for this is the computation of the dynamic weights involves domain information (like the track segment), whereas the fixed weight is an empirical value calculated based on trail runs. Including contextual information has been found effective in different machine learning applications of data-driven anomaly detection \cite{biswas2016approach}, face recognition \cite{davis2005towards}, speech recognition, and query classification \cite{cao2009context}. In our setup, the use of contextual information in the reward function helps us compute a superior dynamic weights compared to fixed weights. We hypothesize these weights are optimal and are required to improve the safety and the performance of the system. Currently these weights are learnt offline, but an online learning technique that adjusts weights on the fly could be useful to tackle new unseen contexts like track, lighting levels, etc.  

% Throughout this work we have emphasized the need for computing dynamic ensemble weights for the weighted simplex strategy. The dynamic weights prov is necessary for systems operating in dynamically changing environments (like changing track segments in our experiments) 

% \textcolor{red}{earlier you mentioned optimal dynamic weights - discuss is needed why is this optimal and how this can be extended for online learning in future work}

\section{Resource Management}
\label{sec:RM}
The dual controller operations of the weighted simplex strategy requires significant computational resources, which are not often available in small scale autonomous vehicles such as those used in hospitals \cite{ozkil2009service}, warehouse \cite{bhasin2016amazon}, and laboratory research testbeds \cite{F1/10}, \cite{karaman2017project}. For the DeepNNCar, the weighted simplex strategy workload increased the power consumption, CPU utilization, and temperature of the RPi3 beyond 70\textdegree C (configured soft limit). Beyond the soft limit, the clock speed and the operating voltage of RPi3 are reduced \cite{RPi}, which could affect the performance of the car. We can address this problem in two ways: (1) \textit{multiple computational units} -- deploy multiple computational devices on-board the DeepNNCar, and distribute the tasks among them. However, this approach requires additional external power sources, which increases development cost of the platform, and (2) \textit{computation offloading} -- as the RPi3 supports WiFi connectivity, we setup wireless communication with other edge devices \footnote{Devices that have similar computational capacity as the onboard RPi3.} or fog devices \footnote{Devices that have higher computational capacity than the onboard RPi3.} and offload some tasks. In this work, we use the computation offloading approach to keep the development costs low, and utilize the wireless communication capability that enables computation on other available devices. 

% An optimal resource management strategy is needed for computation offloading at runtime.

% For performing computation offloading at runtime an optimal resource management strategy is needed.  

\begin{figure}[t]
\centering
\setlength{\belowcaptionskip}{-10pt}
 \includegraphics[width=\columnwidth]{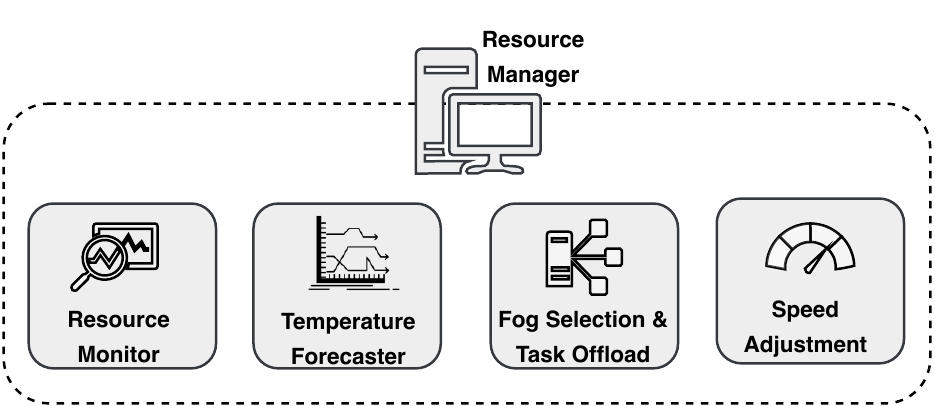}
 \caption{Resource Manager is responsible for resource monitoring, temperature forecasting, fog selection and task offload, and speed adjustment.}
 \label{fig:RM}
%  \vspace{-0.4em}
 \end{figure}

For this, we designed a Resource Manager (RM) that performs the following tasks: (1) resource monitoring -- continuous monitoring of resource state (temperature and CPU Utilization), (2) temperature forecasting-- forecast the temperature of the RPi3 based on current temperature and CPU utilization, (3) fog device selection and task offloading -- selecting an optimal fog device and offloading non-critical tasks, and (4) vehicle speed adjustment-- adjusting the vehicle speed ($\nu$) according to variations in the inference pipeline times. The workflow of the RM is illustrated in \cref{fig:RM}.

\subsection{Resource Monitoring}
For resource constrained computation platforms (e.g. RPi and NVIDIA Jetson), it is important to continuously monitor system utilization including CPU, memory, network, and disk. Variations in these utilization levels could result in degraded computational performance of the system. In our case, temperature and CPU utilization of RPi3 are identified to be the important system parameters that must be monitored, as they directly affect both the computational and hardware performance of the DeepNNCar. 

To perform resource monitoring, the RM in the background continuously monitors the temperature and CPU utilization of the RPi3. This monitoring capability gives the ability to enact preventive measures that could avert the RPi3 from overheating. We use the psutil \cite{rodolapsutil} library for retrieving utilization information and it runs in the background once every 30 seconds. 

\subsection{Temperature Forecasting}
Forecasting the future system utilization based on the current workload is important for resource constrained computation platforms. The forecasting model when used in conjunction with the resource monitor can aid resource management by allowing time to enact mitigation strategies. For example, in our experiments, forecasting the RPi3 temperature based on current workload helped in preparing for task offloading (see \cref{fig:temperatureOffloading}).

The RM is responsible for forecasting the temperature of the RPi3 based on the current temperature and the CPU utilization (which represents the workload). To perform this forecasting, we designed a simple 2-layer DNN with 20 and 40 neurons in the first and second layer respectively. To collect data, we set up an experiment in which we continuously monitored temperature, CPU utilization, and offload status (indicated if the task was performed on the RPi3 or offloaded to a fog device). 

The dataset consisted off three hours of resource monitoring data and was a representation of the entire offload process. We trained the model for 200 epochs with the data collected from the system utilization experiment. The performance and the accuracy of the model is discussed in \cref{sec:evaluation} and illustrated in \cref{fig:NNtemperatureOffloading}. The trained model was then deployed at runtime along with the resource monitor to forecast the temperature of the RPi3 in the next 30 seconds. 

% \begin{table}[t]
%     \centering
%     \setlength{\belowcaptionskip}{-6pt}
%     \renewcommand{\arraystretch}{1.2}
%       \footnotesize
%     \begin{tabular}{|p{1.35cm}|p{1.35cm}|p{1.35cm}|p{1.35cm}|p{1cm}|}
%     \hline
%      Fog 1 \newline Latency(ms) & Fog 2\newline Latency(ms) & Fog 1 Avg \newline Latency(ms) & Fog 2 Avg \newline Latency(ms) & Selected \newline Fog Device\\
%     \hline
%     13.0 \newline 11.4 \newline 11.2 & 10.8 \newline 11.7 \newline 14.0 & 11.87 & 12.17 & Fog 1\\
%      \hline
%     11.7 \newline 11.9 \newline 10.1 & 11.4 \newline 10.6 \newline 10.3 & 11.23 & 10.77 & Fog 2\\    
%     \hline
%     \end{tabular}
%     \caption{The experimental testbed for the fog device selection experiments had Fog 1 device (laptop), Fog 2 device (desktop), and each of these were connected to different WiFi networks. The results show how the selection of a fog device is performed by computing the average of three latency pings. Then, the fog device with the lowest latency is selected for offloading.}
%     \label{Table:FogTable}
%     % \vspace{-0.1in}
% \end{table}

% \usepackage{graphicx}
% \usepackage{graphicx}
\begin{table}[]
\centering
\footnotesize
\begin{tabular}{|c|c|c|c|c|}
\hline
\begin{tabular}[c]{@{}c@{}}Fog1\\ Latency \\ (ms)\end{tabular} & \multicolumn{1}{l|}{\begin{tabular}[c]{@{}c@{}}Fog2\\      Latency \\ (ms)\end{tabular}} & \begin{tabular}[c]{@{}c@{}}Fog1\\ avg Latency\\ (ms)\end{tabular} & \begin{tabular}[c]{@{}c@{}}Fog2\\ avg Latency\\ (ms)\end{tabular} & \begin{tabular}[c]{@{}c@{}}Selected \\ Fog\\ Device\end{tabular} \\ \hline
\begin{tabular}[c]{@{}c@{}}13.0\\ 11.4\\ 11.2\\\end{tabular}        & \begin{tabular}[c]{@{}c@{}}10.8\\ 11.7\\ 14.0\\\end{tabular}                                  & 11.87                                                                      & 12.17                                                                      & Fog1                                                             \\ \hline
\begin{tabular}[c]{@{}c@{}}11.7\\ 11.9\\ 10.1\\\end{tabular}        & \begin{tabular}[c]{@{}c@{}}11.4\\ 10.6\\ 10.3\\\end{tabular}                                  & 11.23                                                                      & 10.77                                                                      & Fog2                                                             \\ \hline
\end{tabular}%
\caption{Fog device selection based on the average of three latency pings. The fog device with the lowest latency is selected for offloading. The experimental testbed for the fog device selection experiments had Fog1 device (laptop), Fog2 device (desktop), and each of these were connected to different WiFi networks.}
\label{Table:FogTable}
\end{table}

\subsection{Fog Device Selection and Task Offloading}
The RM continuously performs a latency ping every 10 seconds using the iPerf \cite{iperf} networking tool to the available fog devices. After 30 seconds, an average time of the latest three ping tests is calculated and the fog device with the lowest average latency is selected for task offloading. \cref{Table:FogTable} shows the latency test results for two 30 second periods, and how we select the appropriate fog device using average latency values. For these experiments we do not perform online discovery, but instead assume that we have prior knowledge of the available fog devices. 

Once, the temperature forecasting model predicts the temperature to exceed 70\textdegree C and an optimal fog device has been selected, the RM prepares to offload the tasks. In our context it is not feasible to transfer the entire task code-base to the fog device at runtime, so we assume these devices have a copy of the code-base pre-installed. During runtime offloading the RM deactivates the code on the RPi3 and activates the same code on the fog device. The message transferring between the RPi3 and the fog device is managed using ZeroMQ (ZMQ). A similar task activation exchange is performed when the predicted temperature falls below 70\textdegree C. The forecasting model and the offloader synchronously work at runtime to plan about the offloading of the computations. 

\subsection{Vehicle Speed Adjustment}
During the task offload process, the inference time $T_R$ (discussed in \cref{sec:middleware}) increases because of the additional network overhead in the wireless communication channel. As the inference time increases, the maximum allowable speed of the vehicle must decrease. For this, the RM instructs the DM to saturate (limit) the top speed of the car. The saturated maximum speed $\nu_{MAX}$ is calculated using the safe distance ($d_S$) which is the closest distance to the track during turning where the car can still safely perform an action to avoid going off the track. The $d_S$ is a track specific quantity which was found to be 0.09m for our track (see \cref{fig:ClientFeedback}). Therefore, any decision taken before reaching this distance will give the car a good turning radius. However, any decision taken after this distance will leave the car with insufficient space to turn and will result in a safety violation. $\nu_{MAX}$ is computed as: $\nu_{MAX} = \frac{d_S}{T_R}$, where $T_R$ is the inference pipeline time. This speed is converted to the corresponding PWM duty cycle value and applied to the RPi3 to control the speed of the car. During task offloading, the Decision Manager Actor (DMA) must wait longer for a reply from the offloaded component due to the latency overhead, which increases the $T_R$.

\section{System Integration}
\label{sec:middleware}

% \textcolor{red}{lets make it a full section}
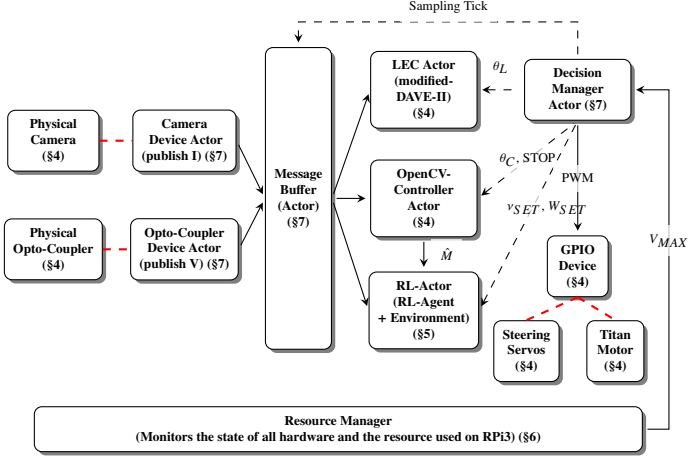
\begin{figure}[t]
\centering
\begin{tikzpicture}[x=1.5cm, y=0.9cm, font=\tiny,
  Component/.style={fill=white, draw, align=center, rounded corners=0.1cm, drop shadow={shadow xshift=0.05cm, shadow yshift=-0.05cm, fill=black}},
  Connection/.style={<->, >=stealth, shorten <=0.06cm, shorten >=0.06cm},
   reqrepl/.style={dashed,->, >=stealth, shorten <=0.06cm, shorten >=0.06cm},
     reqreplrev/.style={dashed,<-, >=stealth, shorten <=0.06cm, shorten >=0.06cm},
       hardware/.style={dashed,->, >=stealth, shorten <=0.06cm, shorten >=0.06cm, gray,ultra thick},
  arr/.style={->, >=stealth, shorten <=0.06cm, shorten >=0.06cm},
  Label/.style={midway, align=center, fill=white, fill opacity=0.75, text opacity=1}
]

\node [Component, minimum width=1.2cm, minimum height=0.75cm] (Camera) at (-2.5, 2.25) {\textbf{Physical}\\ \textbf{Camera}\\ \textbf{(\S \ref{sec:deepnncar})}};

\node [Component, minimum width=1.2cm, minimum height=0.55cm] (Opto) at (-2.5, 0.65) {\textbf{Physical}\\ \textbf{Opto-Coupler}\\ \textbf{(\S \ref{sec:deepnncar})}};

\node [Component, minimum width=1.35cm, minimum height=0.75cm] (CameraDevice) at (-1.35, 2.25) {\textbf{Camera}\\\textbf{ Device Actor}\\ \textbf{(publish I) (\S \ref{sec:middleware})}};

\node [Component, minimum width=1.35cm, minimum height=0.75cm] (OptoCoupler) at (-1.35, 0.65) {\textbf{Opto-Coupler}\\ \textbf{Device Actor}\\ \textbf{(publish V) (\S \ref{sec:middleware})}};

\node [Component, minimum width=0.6cm, minimum height=4cm] (MB) at (-0.35, 1.40) {\textbf{Message}\\ \textbf{Buffer}\\ \textbf{(Actor)}\\ \textbf{(\S \ref{sec:middleware})}};

\node [Component, minimum width=1.4cm, minimum height=0.75cm] (LEC) at (0.75, 3) {\textbf{LEC Actor}\\ \textbf{(modified-}\\ \textbf{DAVE-II)}\\ \textbf{(\S \ref{sec:deepnncar})}};

\node [Component, minimum width=1.4cm, minimum height=0.75cm] (SS) at (0.75, 1.40) {\textbf{OpenCV-}\\ \textbf {Controller}\\ \textbf{Actor}\\ \textbf{(\S \ref{sec:deepnncar})}};

\node [Component, minimum width=1.4cm, minimum height=0.90cm] (RL) at (0.75, -0.25) {\textbf{RL-Actor} \\\textbf{(RL-Agent}\\ \textbf{+ Environment)}\\ \textbf{(\S \ref{sec:Simplex Strategies})}};

\node [Component, minimum width=1.4cm, minimum height=0.75cm] (DMA) at (2.1, 3) {\textbf{Decision}\\ \textbf{Manager}\\ \textbf{Actor (\S \ref{sec:middleware})} };

\node [Component, minimum width=0.8cm, minimum height=0.8cm] (GPIO) at (2.1, 0.40) {\textbf{GPIO}\\ \textbf{Device}\\ \textbf{(\S \ref{sec:deepnncar})}};

\node [Component, minimum width=0.75cm, minimum height=0.75cm] (Servos) at (1.65, -0.85) {\textbf{Steering}\\ \textbf{Servos}\\ \textbf{(\S \ref{sec:deepnncar})}};

\node [Component, minimum width=0.75cm, minimum height=0.75cm] (TI) at (2.43, -0.85) {\textbf{Titan}\\ \textbf{Motor} \\ \textbf{(\S \ref{sec:deepnncar})}};

\node [Component, align=center, minimum width=8cm] (RM) at (0, -2) {\textbf{Resource Manager}\\ \textbf{(Monitors the state of all hardware and the resource used on RPi3) (\S \ref{sec:RM})}};

 \draw[red,thick,dashed]  (Camera.east) to (CameraDevice.west) ;
 \draw[red,thick,dashed]  (Opto.east) to (OptoCoupler.west) ;
 \draw[arr]  (CameraDevice.east) to (MB.west);
\draw[arr]  (OptoCoupler.east) to  (MB.west) ;
 \draw[arr]  (MB.east) to (LEC.west) ;
 \draw[arr]  (MB.east) to (SS.west) ;
 \draw[arr]  (MB.east) to (RL.west) ;
\draw[reqreplrev]  (LEC.east) to node [Label,yshift=0.3cm] {$\theta_L$} (DMA.west) ;
 \draw[reqreplrev]  (SS.east) to node [Label] {$\theta_C$, STOP} (DMA.south) ;
 \draw[arr]  (SS.south) to node [Label, xshift=0.3cm]{$\hat M$} (RL.north) ;
 \draw[reqreplrev]  (RL.east) to node [Label, xshift=0.25cm, yshift=0.1cm] {$\nu_{SET}$, $W_{SET}$} (DMA.south) ;
 \draw[arr]  (DMA.south) to node [Label] {PWM} (GPIO.north) ;

 \draw[red,thick,dashed]  (GPIO.south) to (Servos.north) ;
 \draw[red,thick,dashed]  (GPIO.south) to (TI.north) ;

\draw [reqrepl] (DMA.north) |- node [Label,xshift=-1.7cm,yshift=0.2cm] {Sampling Tick}(0.7,4) -| (MB.north);

 \draw [arr] (RM.east) -| node [Label, yshift=2.5cm] {$V_{MAX}$}(2.9,2) |- (DMA.east);

\end{tikzpicture}
    \caption{A block diagram of DeepNNCar along with actors. There are asynchronous interactions among various actors and thus different messaging patterns were used. The request-reply communications are shown with dotted lines, the publish-subscribe communications are shown in solid lines, and the red dotted lines indicate the hardware connections. Also, refer to the listed section numbers for a detailed description about the components.}
    \label{fig:block}
    \vspace{-0.8em}
\end{figure}

% \textcolor{red}{Ensure these section numbers are correct. Explain that section numbers are shown where the actor description can be found}

% \textcolor{red}{we need to say that we use an actor based design pattern for integration - you can cite the old agha paper on actor based design. Then dont use the term components - use the term actor all over the place. }

We use an actor based design \cite{agha1985actors} to integrate the components discussed in the aforementioned sections. As discussed in \cite{jang2005actor}, an actor is a self-constrained and restartable process that has its own execution thread and communicates synchronously or asynchronously with other actors. The actors of DeepNNCar and the data flow between them is shown in \cref{fig:block}. Communication between actors is done with various ZMQ messaging patterns. The camera provides new images at 30 Hz and the IR opto-coupler continuously collects RPM data to compute the speed of the car. Then, the camera device actor\footnote{A device actor converts hardware sensor information into topics that can be published and subscribed to (see \cite{device2018}).} and the opto-coupler device actor periodically publish the images and current-speed to all subscriber actors. The LEC controller actor, the OpenCV controller actor, and the RL-Actor are aperiodic consumers (see \cite{dubey2011model}) which do not consume the sensor values until prompted by the DMA. 

The interactions between the periodic publishers and aperiodic consumers are handled with the help of a Message Buffer Actor (MBA), which has a one buffer queue to store the published data (both images and current-speed) along with a sequence label. The data in the MBA gets updated according to the sampling period of the sensors. However, this data cannot be published until the MBA has received a sampling tick and a request from the DMA to publish the data of a certain label. Once the MBA receives this request, it publishes the image and current-speed messages to all subscribed actors. Using this data, the LEC actor predicts $\theta_L$, the OpenCV controller computes $\theta_{C}$, track position $\hat M$, and $STOP$ (a command issued if the car goes out of the track), and the RL-Actor computes $W_L$, $W_C$ and $\nu_{SET}$. 

\ul{Decision Manager Actor (DMA)}: This is the key component in the architecture that controls and initiates the entire message exchange process every inference cycle. The DMA issues requests for the sequence label and data $\theta_L$, $\theta_{C}$ from the controllers, and for $W_L$, $W_{C}$, and $\nu_{SET}$ from the RL-Actor. Once the controller and the RL-Actor have finished computation, they respond to the DMA with their label and values. The DMA then matches the labels and computes $\theta_{D}$ using \cref{Eqn:DeepNNCar} before feeding $\theta_{D}$ and $\nu_{SET}$ to the GPIO device actor, which controls the two motors.

After applying the controls to the GPIO, the DMA starts a new cycle. This cycle continues indefinitely until it is terminated by the STOP signal from the OpenCV controller or manually by the user. The tasks performed between two sampling ticks of the DMA is one control cycle of the system and the time taken to perform one control cycle is referred to as the inference pipeline time $T_R$. $T_R$ varies for every control cycle, but the average inference time for the dynamic-weighted simplex strategy is experimentally found to be 130 ms (in \cref{fig:inference}).

\section{Evaluation}
\label{sec:evaluation}
% \textcolor{red}{reposition figures so that they are right next to where they are being discussed. right now figures are all over the place}

We evaluate the performance of the middleware framework and the introduced weighted simplex strategies by deploying it on the DeepNNCar. For this evaluation, we built three different indoor tracks shown in \cref{fig:ClientFeedback}. These tracks were built in our laboratory using 10' x 12' tarps under controlled lighting conditions (higher lighting intensities create reflections on the tarp, causing the LD algorithm to fail). Each track had different geometric shapes and turns. These tracks also look wrinkled since they are folded for storage after experiments are complete. All the experiments were performed using the tracks in the same wrinkled conditions. The LEC was trained on the images collected from Track 1 and Track 2 and then tested on Track 3 to ensure the trained CNN had not overfit (generalized with the training data).

For evaluating the performance of the RM, we designed the DeepNNCar to communicate with a laptop (with an Intel 4-core processor) and a desktop (with AMD Ryzen Threadripper 16-core processor) using wireless communication over WiFi.

\begin{figure}[t]
% \vspace{-0.1in}
    \centering
\begin{tikzpicture}
\begin{axis}[
  ybar,
  area legend,
    font=\footnotesize,
  ytick={0,3,6,9,12,15},
  bar width=4pt,
  width=\columnwidth,
  grid=major,
  height=0.4\columnwidth,
  ylabel style={text width=1.75cm}, ylabel=Out-of-track\\ Occurrences\\,
  xlabel=speed in m/s,
  legend style={at={(0.5,-0.5)},anchor=north,legend columns=-1},
  symbolic x coords=
    {0.25, 0.35, 0.45, 0.55, 0.65},
  xtick=data,
  ]
   \addplot coordinates {(0.25,2)(0.35,3)(0.45,10)(0.55,12)(0.65,13)};
  \addlegendentry{CV};
     \addplot coordinates {(0.25,1)(0.35,3)(0.45,4)(0.55,7)(0.65,12)};
     \addlegendentry{LEC};
      \addplot coordinates {(0.25,0)(0.35,0)(0.45,1)(0.55,3)(0.65,7)};
      \addlegendentry{FW};
         \addplot coordinates {(0.25,0)(0.35,0)(0.45,0)(0.55,1)(0.65,3)};
         \addlegendentry{DW};

\end{axis}
\end{tikzpicture}
  \caption{The out-of-track occurrences by different controllers for different speeds. From figure, CV: driving only with the OpenCV controller, LEC: driving only with the modified Dave-II model, FW: driving with fixed-weighted simplex strategy, and DW: driving with dynamic-weighted simplex strategy. The horizontal axis shows the different speeds of the car during the experiment. The DeepNNCar performs fewer out-of-track occurrences when combined with the dynamic-weighted simplex strategy. The data for this experiment was collected by running DeepNNCar with each of these controllers independently around Track 1 (see \cref{fig:ClientFeedback}) for 10 laps.}
    \label{fig:carsimples}
    % \vspace{-0.1in}
\end{figure}
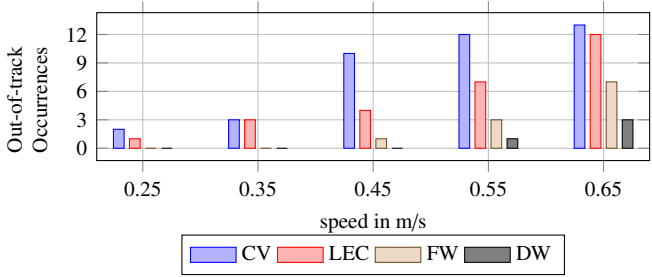

\subsection{Experiments on Weighted Simplex Performance}

\ul{Out-of-track Occurrences}: To evaluate the performance of the LEC controller, the OpenCV controller, and the weighted simplex strategies, we deployed them on DeepNNCar and performed various trial runs. \cref{fig:carsimples} shows the number of out-of-track occurrences performed by the different controllers at different speeds. We varied the speed between [0.25 - 0.65] m/s and ran the different controllers separately for 10 laps around Track 1. From \cref{fig:carsimples} it is evident that, at low speeds (< 0.35) m/s, both variants of weighted simplex strategy have lower out-of-track occurrences compared to the independent controllers. At higher speeds (0.65 m/s), the dynamic-weighted simplex strategy outperforms all other controllers reducing the number of out-of-track occurrences to 3. 

\begin{figure}[h!]
    \centering
    \begin{tikzpicture}
\begin{axis}[
      width=\columnwidth,
      height=0.23\textwidth,
      font=\footnotesize,
      grid=major,
      JobOfferStyle/.style={boxplot={box extend=0.5,}, blue, solid, fill=blue!20, mark=x},
      ytick={1, 2, 3, 4},
      yticklabels={{CV}, {LEC}, {FW}, {DW}, {(e)}, {(f)},{(g)}},
       ResourceOfferStyle/.style={boxplot={box extend=0.5,}, brown, solid, fill=brown!20, mark=x},
      RegisterStyle/.style={boxplot={box extend=0.5,}, blue, solid, fill=blue!20, mark=x},
      MediatorStyle/.style={boxplot={box extend=0.5,}, green, solid, fill=green!20, mark=x},
      OfferMatchTimeStyle/.style={boxplot={box extend=0.5,}, red, solid, fill=red!20, mark=x},
      ResultTimeStyle/.style={boxplot={box extend=0.5,}, gray, solid, fill=gray!20, mark=x},
      ExecutionTimeStyle/.style={boxplot={box extend=0.5,}, orange, solid, fill=orange!20, mark=x},
      xlabel={Speeds across different strategies in m/s }
    ]
    \addplot+[MediatorStyle, boxplot={draw position=1}] table [col sep=comma, y=SS] {results/speed.csv};
     \addplot+[ResourceOfferStyle, boxplot={draw position=2}] table [col sep=comma, y=CNN] {results/speed.csv};
     \addplot+[OfferMatchTimeStyle, boxplot={draw position=3}] table [col sep=comma, y=AMSimplex] {results/speed.csv};
     \addplot+[RegisterStyle, boxplot={draw position=4}] table [col sep=comma, y=RLSimplex] {results/speed.csv};
\end{axis}
\end{tikzpicture}
  \caption{Speeds in meters per second  of CV: driving only with the OpenCV controller, LEC: driving only with the modified Dave-II model, FW: driving with fixed-weighted simplex strategy, and (d) DW: driving with dynamic-weighted simplex strategy.}
    \label{fig:speed}
\end{figure}
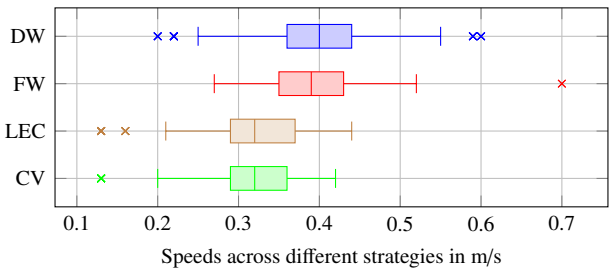

\ul{Speed Performance}: \cref{fig:speed} shows the maximum speed (represented as anomalies in the figure) and the average speed performance of the different controllers. To collect data for this experiment we ran the car with different controllers independently for 10 laps around Track 1. Both the OpenCV controller and the LEC controller operate with approximately the same average speeds of 0.29 - 0.39 m/s. This is because these controllers have a small number of out-of-track occurrences (see \cref{fig:carsimples}) in this speed range. However, the two weighted simplex strategies operate in a higher speed range because they perform significantly less out-of-track occurrences for speeds (< 0.4) m/s. Notably, the dynamic-weighted simplex strategy operates at a higher speed with lower out-of-track occurrences as compared to the other controllers.  

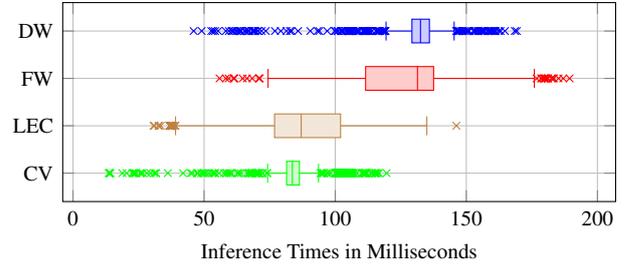
\begin{figure}[h!]
\begin{tikzpicture}
\begin{axis}[
      width=\columnwidth,
      height=0.23\textwidth,
      font=\footnotesize,
      grid=major,
      xtick={0,50,100,150,200},
      JobOfferStyle/.style={boxplot={box extend=0.5,}, blue, solid, fill=blue!20, mark=x},
      ytick={1, 2, 3, 4},
      yticklabels={{CV}, {LEC}, {FW}, {DW}, {(e)}, {(f)},{(g)}},
       ResourceOfferStyle/.style={boxplot={box extend=0.5,}, brown, solid, fill=brown!20, mark=x},
      RegisterStyle/.style={boxplot={box extend=0.5,}, blue, solid, fill=blue!20, mark=x},
      MediatorStyle/.style={boxplot={box extend=0.5,}, green, solid, fill=green!20, mark=x},
      OfferMatchTimeStyle/.style={boxplot={box extend=0.5,}, red, solid, fill=red!20, mark=x},
      ResultTimeStyle/.style={boxplot={box extend=0.5,}, gray, solid, fill=gray!20, mark=x},
      ExecutionTimeStyle/.style={boxplot={box extend=0.5,}, orange, solid, fill=orange!20, mark=x},
      xlabel={Inference Times in Milliseconds}
    ]
    \addplot+[MediatorStyle, boxplot={draw position=1}] table [col sep=comma, y=SS] {results/pipeline.csv};
     \addplot+[ResourceOfferStyle, boxplot={draw position=2}] table [col sep=comma, y=CNN] {results/pipeline.csv};
     \addplot+[OfferMatchTimeStyle, boxplot={draw position=3}] table [col sep=comma, y=AMSimplex] {results/pipeline.csv};
     \addplot+[RegisterStyle, boxplot={draw position=4}] table [col sep=comma, y=RLSimplex] {results/pipeline.csv};
\end{axis}
\end{tikzpicture}
  \caption{Inference times in milliseconds of CV: driving only with the OpenCV controller, LEC: driving only with the modified Dave-II model, FW: driving with fixed-weighted simplex strategy, and (d) DW: driving with dynamic-weighted simplex strategy. Inference time is affected by the state (computational load, temperature), so we have very high variance in the inference times.}
    \label{fig:inference}
\end{figure}

\ul{Inference Times}: We also evaluate the performance of the controllers based on inference times. \cref{fig:inference} illustrates the inference times of the different controllers. It is evident from the figure that, the dual controller operation and the decision computation steps of the weighted simplex strategy increases the inference times of the car. The dynamic-weighted simplex strategy has an average inference time of 130 ms, and the fixed-weighted simplex strategy has an average of 120 ms whereas both independent controllers have smaller inference times of about 80 ms. The higher number of anomalies in this graph is
due to the variation in inference times based on the computational load and the onboard temperature of the RPi3. As the RPi3 gets overheated and overloaded, the chances of varying inference times is higher.

% \begin{table}[h!]
% \setlength{\tabcolsep}{3pt}
% \centering
%  \footnotesize
%     \setlength{\belowcaptionskip}{-2pt}
%     \begin{tabular}{|p{2.5cm}|p{1.5cm}|p{1.5cm}|p{1.4cm}|}
%     \hline
%      Simplex Strategy & Straight Segment & Near Curved Segment & In Curved Segment\\
%     \hline
%     Dynamic Weights \newline ($W_L$, $W_C$) & (0.95, 0.05) & (0.85, 0.15) & (0.8, 0.2)\\
%     \hline
%      Fixed Weights \newline ($W_L$, $W_C$) & (0.8, 0.2) & (0.8, 0.2) & (0.8, 0.2)\\
%     \hline
%     \end{tabular}
%     \caption{Comparing the ensemble weights of different simplex strategies. For the dynamic-weighted simplex strategy the weights were dynamically updated by the Q-learning algorithm. For the fixed-weighted simplex strategy we have a fixed weight for all the track segments, these weights were manually tuned by a human supervisor.}
%     \label{Table:RLWeights}
%     \vspace{-0.1in}
% \end{table}

% \usepackage{graphicx}
\begin{table}[h!]
\centering
\footnotesize
\begin{tabular}{|c|l|c|c|}
\hline
\begin{tabular}[c]{@{}c@{}}Weighted\\ Simplex Strategy\end{tabular} & \begin{tabular}[c]{@{}l@{}}Straight \\ Segment\end{tabular} & \begin{tabular}[c]{@{}c@{}}Near\\ Curved\\ Segment\end{tabular} & \begin{tabular}[c]{@{}c@{}}In\\ Curved\\ Segment\end{tabular} \\ \hline
\begin{tabular}[c]{@{}c@{}}Dynamic Weights\\ (WL, WC)\end{tabular}  & \multicolumn{1}{c|}{0.95, 0.05}                             & 0.85, 0.15                                                      & 0.80, 0.20                                                    \\ \hline
\begin{tabular}[c]{@{}c@{}}Fixed Weights\\ (WL, WC)\end{tabular}    & 0.80, 0.20                                                  & \multicolumn{1}{l|}{0.80, 0.20}                                 & 0.80, 0.20                                                    \\ \hline
\end{tabular}%
\caption{Comparing the ensemble weights of different simplex strategies. For the dynamic-weighted simplex strategy the weights were dynamically updated by the Q-learning algorithm. For the fixed-weighted simplex strategy we have a fixed weight for all the track segments, these weights were manually tuned by a human supervisor.}
    \label{Table:RLWeights}
\end{table}

% \textcolor{red}{make sure fonts in all tables are footnote size}

\ul{Fixed and Dynamic Weights}: We performed an experiment to show how the weights of different weighted simplex strategies vary across different segments of the track. For this we clustered the Track 1 into three segments: Straight, Near Curved, and In Curved. Then as the car ran using the two weighted simplex strategies, we recorded the ensemble weights. We later classified the weights into the three segments. From \cref{Table:RLWeights} we see the weights used in the fixed-weighted strategy, remains constant across the three track segments. However, in the dynamic-weighted strategy the weights change dynamically and are different for the three different track segments. The dynamic-weights in each segment are not always the same as shown in \cref{Table:RLWeights}, but they vary $\pm$ 0.05. To simplify the table we have just listed a single value that occurred most often for a particular track segment.

% In \cref{Table:RLWeights}, for the dynamic-weighted simplex strategy we have listed a single value that occurred most often for a particular track segment. 

\ul{Summary}: \cref{fig:carsimples} shows the introduced dynamic-weighted simplex strategy can operate at higher speeds with lower out-of-track occurrences. However, this comes with a penalty of increased inference times (see \cref{fig:inference}). From these results, it can be inferred that dynamic blending of the simplex weights helps in reducing soft constraint violations while achieving higher performance.

\subsection{Experiments on Task Offloading}

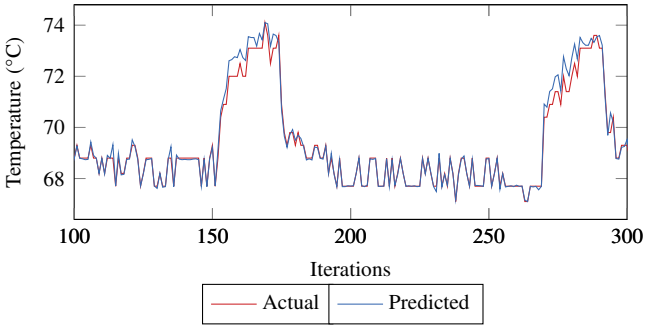
\begin{figure}[h!]
\centering
\begin{tikzpicture}
\begin{axis}[
  font=\footnotesize,
  width=\columnwidth,
  height=0.5\columnwidth,
  ylabel style={text width=2cm}, ylabel=Temperature (\textdegree C),
  xlabel=Iterations,
  legend style={at={(0.35,-0.30)},
      anchor=north,legend columns=-1},
   xmin=100,
  xmax=300,
    grid style={line width=.1pt, draw=gray!10},
    major grid style={line width=.2pt,draw=gray!50},
    ]
\addplot[solid, redLine] table[col sep=comma,x=sequence,y=Actual]{results/validation2.csv};
\addlegendentry{Actual}
 \end{axis}
  \begin{axis}[
  font=\footnotesize,
  width=\columnwidth,
  height=0.5\columnwidth,
  legend style={at={(0.6,-0.30)},
      anchor=north,legend columns=-1},
     xmin=100,
  xmax=300,
  axis y line=none,
 ]
 \addplot[solid,blueLine] table[col sep=comma,x=sequence,y=Predicted]{results/validation2.csv};
\addlegendentry{Predicted}
 \end{axis}
\end{tikzpicture}
\caption{The result of the DNN temperature predictions vs. the actual temperature values. The DNN forecasts the temperature of the computing unit based on the current state (temperature, CPU utilization). We see the DNN predictions closely match the actual values. The graph shows a subset of iterations (total 10000 iterations).}
\label{fig:NNtemperatureOffloading}
\end{figure}

% \vspace{0.1in}

% \begin{figure}
% \begin{tikzpicture}
% \begin{axis}[
%   font=\footnotesize,
%   width=\columnwidth,
%   height=0.5\columnwidth,
%   ylabel style={text width=2cm}, ylabel=Temperature (\textdegree C),
%   xlabel=Iterations,
%   xmin=1400,
%   xmax=1700,
%     grid style={line width=.1pt, draw=gray!10},
%     major grid style={line width=.2pt,draw=gray!50},
%     ]
% \addplot[solid, blueLine] table[col sep=comma,x=Sequence,y=Temperature] {results/temp.csv};
%  \end{axis}
%   \begin{axis}[
%   font=\footnotesize,
%   width=\columnwidth,
%   height=0.5\columnwidth,
%      xmin=1400,
%   xmax=1700,
%   ytick={0,1},
%       yticklabels={{(off)}, {(on)}},
%   axis y line*=right,
%   legend pos=north west,
%  ]
%  \addplot[solid,redLine] table[col sep=comma,x=Sequence,y=Offloading] {results/temp.csv};
%  \end{axis}
% \end{tikzpicture}
% \setlength{\abovecaptionskip}{-4pt}
% \setlength{\belowcaptionskip}{-10pt}
% \caption{The effect of offloading the tasks in response to high temperature per iteration of the inference pipeline. The trigger to offload the task is 70\textdegree C. The blue line shows the temperature in Celsius. The red line shows when the tasks were offloaded to the fog (on=on fog, off=off fog). The graph shows a subset of iterations (total 10000 iterations).}
% \label{fig:temperatureOffloading}
% % \end{figure}
% \vspace{-0.4em}
% \end{figure} 

\ul{Temperature variations}: As discussed in \cref{sec:RM}, we design a temperature forecasting model using a DNN. \cref{fig:NNtemperatureOffloading} shows the predicted  DNN temperatures vs. the actual temperature readings. We can see the DNN predictions closely match to the actual ones. To quantify the performance of the temperature forecaster, we use Mean Absolute Percentage Error (MAPE) as the metric. The results of MAPE = 0.16\% across 1000 predictions indicates the model accurately predicts the temperature of the RPi3. 

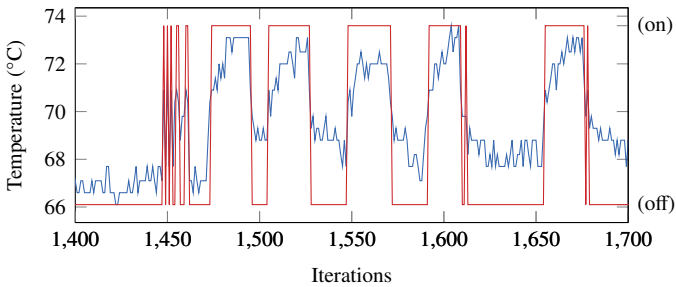
\begin{figure}[h!]
\begin{tikzpicture}
\begin{axis}[
  font=\footnotesize,
  width=\columnwidth,
  height=0.5\columnwidth,
  ylabel style={text width=2cm}, ylabel=Temperature (\textdegree C),
  xlabel=Iterations,
   xmin=1400,
  xmax=1700,
    grid style={line width=.1pt, draw=gray!10},
    major grid style={line width=.2pt,draw=gray!50},
    ]
\addplot[solid, blueLine] table[col sep=comma,x=Sequence,y=Temperature] {results/temp.csv};
 \end{axis}
  \begin{axis}[
  font=\footnotesize,
  width=\columnwidth,
  height=0.5\columnwidth,
     xmin=1400,
  xmax=1700,
   ytick={0,1},
      yticklabels={{(off)}, {(on)}},
  axis y line*=right,
  legend pos=north west,
 ]
 \addplot[solid,redLine] table[col sep=comma,x=Sequence,y=Offloading] {results/temp.csv};
 \end{axis}
\end{tikzpicture}
\setlength{\abovecaptionskip}{-4pt}
\caption{The effect of offloading the tasks in response to high temperature per iteration of the inference pipeline. The trigger to offload the task is 70\textdegree C. The blue line shows the temperature in Celsius. The red line shows when the tasks were offloaded to the fog (on=on fog, off=off fog). The graph shows a subset of iterations (total 10000 iterations).}
\label{fig:temperatureOffloading}
% \end{figure}
\vspace{-0.4em}
\end{figure}

\ul{Task Offloading}: To evaluate the performance of the task offloader, we continuously offloaded the task onboard the DeepNNCar to any one of the available fog devices (laptop or a desktop). We then note the temperature variations when the tasks stay onboard and when they are offloaded. The fluctuating blue lines in \cref{fig:temperatureOffloading} shows the temperature variations when the task was moved on and off the RPi3. The red lines indicate if the tasks were executed on or off the RPi3. As observed, the temperature drops below the threshold (70\textdegree C) when the tasks get offloaded. 

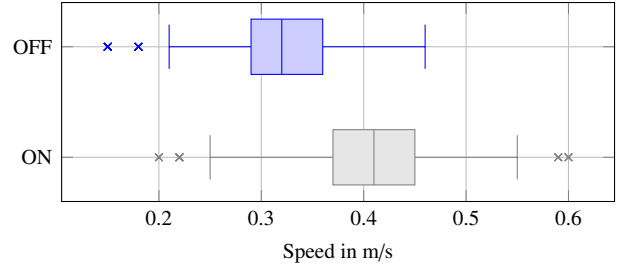
\begin{figure}[h!]
\centering
\begin{tikzpicture}
\begin{axis}[
      width=\columnwidth,
      height=0.23\textwidth,
      font=\footnotesize,
      grid=major,
      JobOfferStyle/.style={boxplot={box extend=0.5,}, blue, solid, fill=blue!20, mark=x},
      ytick={1, 2},
      yticklabels={{ON}, {OFF}, {(c)}, {(d)}, {(e)}, {(f)},{(g)}},
       ResourceOfferStyle/.style={boxplot={box extend=0.5,}, brown, solid, fill=brown!20, mark=x},
      RegisterStyle/.style={boxplot={box extend=0.5,}, black, solid, fill=blue!20, mark=x},
      MediatorStyle/.style={boxplot={box extend=0.5,}, green, solid, fill=green!20, mark=x},
      OfferMatchTimeStyle/.style={boxplot={box extend=0.5,}, red, solid, fill=red!20, mark=x},
      ResultTimeStyle/.style={boxplot={box extend=0.5,}, gray, solid, fill=gray!20, mark=x},
      ExecutionTimeStyle/.style={boxplot={box extend=0.5,}, orange, solid, fill=orange!20, mark=x},
      xlabel={Speed in m/s}
   ]
    \addplot+[ResultTimeStyle, boxplot={draw position=1}] table [col sep=comma, y=onboard] {results/fog-edge-speed.csv};
     \addplot+[JobOfferStyle, boxplot={draw position=2}] table [col sep=comma, y=offloaded] {results/fog-edge-speed.csv};
    \end{axis}
\end{tikzpicture}
    \caption{Speed readjustment during offload (m/s) ON: dynamic-weighted simplex strategy with all tasks executed onboard, and OFF: dynamic-weighted simplex strategy with RL task offloaded (Q-table was offloaded).}
    \label{fig:offloadspeed}
\end{figure}

\ul{Speed Performance}: \cref{fig:offloadspeed} shows that the DeepNNCar can operate at higher speeds ($\approx$ 0.42 m/s) when all tasks are preformed onboard. As tasks are offloaded to fog devices, the operating speed decreases to 0.34 m/s. This is because of the continuous speed adjustment performed to compensate for the increased inference times (due to increased latency overhead).

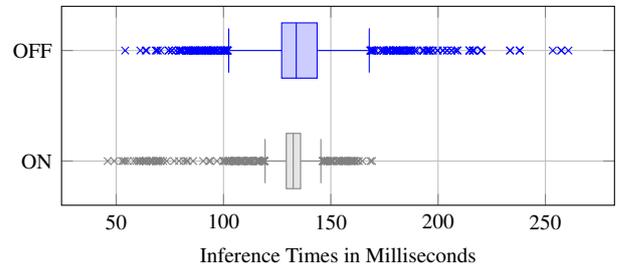
\begin{figure}[h!]
\begin{tikzpicture}
\begin{axis}[
      width=\columnwidth,
      height=0.23\textwidth,
      font=\footnotesize,
      grid=major,
      JobOfferStyle/.style={boxplot={box extend=0.5,}, blue, solid, fill=blue!20, mark=x},
      ytick={1, 2, 3, 4},
      yticklabels={{ON}, {OFF}, {(c)}, {(d)}, {(e)}, {(f)},{(g)}},
       ResourceOfferStyle/.style={boxplot={box extend=0.5,}, brown, solid, fill=brown!20, mark=x},
      RegisterStyle/.style={boxplot={box extend=0.5,}, black, solid, fill=blue!20, mark=x},
      MediatorStyle/.style={boxplot={box extend=0.5,}, green, solid, fill=green!20, mark=x},
      OfferMatchTimeStyle/.style={boxplot={box extend=0.5,}, red, solid, fill=red!20, mark=x},
      ResultTimeStyle/.style={boxplot={box extend=0.5,}, gray, solid, fill=gray!20, mark=x},
      ExecutionTimeStyle/.style={boxplot={box extend=0.5,}, orange, solid, fill=orange!20, mark=x},
      xlabel={Inference Times in Milliseconds}
    ]
    \addplot+[ResultTimeStyle, boxplot={draw position=1}] table [col sep=comma, y=onboard] {results/fog-edge-pipeline.csv};
     \addplot+[JobOfferStyle, boxplot={draw position=2}] table [col sep=comma, y=oflloaded] {results/fog-edge-pipeline.csv};
    \end{axis}
\end{tikzpicture}
\setlength{\belowcaptionskip}{-10pt}
    \caption{ON: dynamic-weighted simplex strategy with all tasks executed onboard, and OFF: dynamic-weighted simplex strategy with RL task offloaded (Q-table was offloaded). The higher overloaded, and overheated the RPi3 gets, the inference times gets higher. Inference time is affected by the state (computational load, temperature), so we have very high variance in the inference times.}
     \label{fig:offload}
    %  \vspace{0.1in}
\end{figure}

\ul{Inference Times}: \cref{fig:temperatureOffloading} shows task offloading performed to maintain the RPi3's temperature below the threshold (70\textdegree C). During offloading, the inference times increase because of the network overhead involved in sending out the computations to the fog devices. The inference time comparison of the car when tasks get offloaded vs. not-offloaded is shown in \cref{fig:offload}. The dynamic-weighted simplex strategy with all tasks performed onboard has a lower inference time ($\approx$ 130 ms) compared to the inference time ($\approx$ 140 ms) with RL tasks offloaded (Q-Table was offloaded). 

The higher number of anomalies in this graph is because inference times vary based on the computational load and the onboard temperature of the RPi3. As the RPi3 overheats and gets overloaded, the chance of varying inference times are higher. In addition, during offloading the wireless exchange of messages is performed over Vanderbilt University WiFi. The network traffic variations in the WiFi adds on to the latency overhead.

\ul{Summary}: It is evident from \cref{fig:temperatureOffloading} that the resource manager tries to keep a tight check on the RPi3 temperature by offloading the RL task onto available fog devices. However, to balance the increase in inference times during offloading (see \cref{fig:offload}), the resource manager has to penalize or adjust the speed of the car. The average speed during offload is 0.34 m/s, compared to 0.42 m/s without the offload. Since, the primary concern of this paper is to reduce the soft constraint violations, the slight penalization in the performance (speed) is acceptable. For small-scale CPS systems with limited computational capacity, we would recommend the use of a resource manager along with an optimal dynamic-weighted blending strategy as discussed in this work.  

\section{Conclusion}
\label{sec:conclusion}
In this work, we have discussed the problems associated in using LECs for perception in autonomous systems. We have further implemented an LEC based end-to-end learning controller on our physical testbed, the DeepNNCar. We have also discussed the two key challenges of using the classical Simplex Architecture. They are: (1) designing effective decision logic, and (2) mitigating sudden transitions. To address these research challenges, we discuss: (1) a \textit{dynamic-weighted simplex strategy} which computes dynamic simplex weights according to the segments of the track, (2) a \textit{middleware framework} which allows the integration of the introduced dynamic-weighted strategy and provides a resource manager for monitoring the onboard computational unit, and (3) a \textit{hardware testbed} to deploy and test the proposed concepts. We have further evaluated the performance of the dynamic-weighted simplex strategy in terms of its capability to reduce soft constraint violations while improving the system's performance. Our results show the dynamic-weighted simplex strategy works well at higher speeds ($\sim$ 0.4 m/s) with low out-of-track occurrences as compared to only the LEC driven controller and the OpenCV controller. In addition, the evaluation results also show the resource manager to be effective in mitigating the computational overload generated by the dynamic-weighted simplex strategy. 

This framework finds utility in factories, warehouses, hospitals where various levels of safe autonomy are required to perform tasks. Currently, the dynamic-weighted simplex strategy is suitable for systems which can tolerate soft constraint violations. However, in the future we would like to extend this strategy to compute dynamic ensemble weights that can be applicable to systems with hard constraints. Further, we would like to extend our middleware framework with an architecture such as CHARIOT \cite{pradhan2015chariot}, which provides a mechanism for achieving autonomous resilience. Network failure is a significant problem in our current setup. Using CHARIOT could help us detect this failure and mitigate it.

\textbf{Acknowledgements}: 
This work was supported by DARPA's Assured Autonomy project and Air Force Research Laboratory. Any opinions, findings, and conclusions or recommendations expressed in this material are those of the author(s) and do not necessarily reflect the views of DARPA or AFRL.

\balance

\section*{References}
\bibliographystyle{elsarticle-num}
\bibliography{references/references.bib,references/bibliography.bib}

\appendix
\renewcommand{\thesubsection}{\Alph{subsection}}
\renewcommand\thefigure{A.\arabic{figure}}
\setcounter{figure}{0}
\renewcommand\thetable{A.\arabic{table}}
\setcounter{table}{0}
\acuseall
\end{document}